\documentclass[11pt]{article}
\usepackage{graphicx} 

\usepackage{natbib}
\usepackage{caption}
\usepackage{siunitx}
\usepackage{ifthen}
\newboolean{isSingleColumn}

\makeatletter
\if@twocolumn
    \setboolean{isSingleColumn}{false}
\else
    \setboolean{isSingleColumn}{true}
\fi
\newcommand{\adaptiveHeader}[1]{%
    \ifthenelse{\boolean{isSingleColumn}}{%
        \paragraph{#1} 
    }{%
        \textbf{#1}    
    }%
}

\usepackage{fullpage}
\usepackage{hyperref}
\usepackage{authblk}
\usepackage{algorithm}
\usepackage{algorithmic}
\usepackage{booktabs}
\usepackage{complexity}
\usepackage{amsmath}
\usepackage{amssymb}
\usepackage{amsthm}
\usepackage{mathabx}
\usepackage{xcolor}
\usepackage{subfigure}
\usepackage{multirow}
\usepackage[capitalize,noabbrev]{cleveref}

\newcommand{\upt}{^{(t)}}
\newcommand{\scope}{\operatorname{scope}}
\DeclareMathOperator*{\argmax}{arg\,max}
\newcommand{\imgheight}{100 pt}
\newcommand{\imgwidth}{100 pt}
\newcommand{\tabwidth}{\linewidth}

\title{Distributed Constraint Optimization via Online Learning and Iterative Pricing with Application to Large-Scale Satellite Scheduling}

\author[1]{Itai Zilberstein\thanks{Equal contribution. Correspondence to \texttt{izilbers@cs.cmu.edu, prajbhan@cs.cmu.edu}}}
\author[1]{Pranav Rajbhandari\protect\footnotemark[1]}
\author[2]{Steve Chien}
\author[1,3]{Tuomas Sandholm}

\affil[1]{Department of Computer Science, Carnegie Mellon University}
\affil[2]{Jet Propulsion Laboratory, California Institute of Technology}
\affil[3]{\small{Additional affiliations: Strategy Robot, Inc., Strategic Machine, Inc., Optimized Markets, Inc.}}

\begin{document}

\maketitle
\thispagestyle{empty}

\begin{abstract}
\textit{Distributed constraint optimization problems (DCOPs)} provide a popular framework for distributed decision making under limited communication, but many real-world instances are too large to solve monolithically. We address this challenge from two complementary directions. We revisit the connection between DCOPs and potential games, and adapt modern online learning algorithms for equilibrium finding to DCOPs. We show that these algorithms are competitive with representative incomplete DCOP algorithms. We then turn to decomposition frameworks for large-scale DCOPs, motivated by large-scale decentralized satellite scheduling. We propose a new framework that separates a DCOP into two interacting subproblems: a high-level meta-DCOP for task allocation, and independent local optimization problems for scheduling. To couple the two levels, we develop a novel iterative pricing method that updates the meta-level utilities using feedback from the local optimizers. Combining our online learning methods with our iterative pricing framework, we obtain near-optimal performance on real-world decentralized satellite scheduling problem instances, fulfilling over \(99\%\) of observation requests compared with \(87\%\) for state-of-the-art baselines. 

\end{abstract}
 
\newpage
\setcounter{page}{1}

\section{Introduction}

\textit{Distributed constraint optimization problems (DCOPs)} model collaborative multiagent systems in which agents coordinate local variable assignments to optimize a shared global objective~\citep{fioretto2018distributed}.
DCOPs provide a popular framework for distributed decision making under limited communication, with applications ranging from mobile sensor teams~\citep{ca-dcop-mst} to satellite scheduling~\citep{zilberstein2025decentralized}.
However, solving DCOPs optimally is \NP-hard~\citep{adopt}.
Complete algorithms typically incur exponential time or communication complexity~\citep{adopt,syncbb,gershman2009asynchronous}.
Practical DCOP deployments often rely on incomplete local-search, sampling, or inference-based methods, including DSA~\citep{dsa}, MGM~\citep{mgm}, GDBA~\citep{gdba}, D-Gibbs~\citep{dgibs}, and Maxsum variants~\citep{maxsumadvp}. While these methods are lightweight, many real-world domains remain too large or structurally complex to model and solve directly as monolithic DCOPs.

An important application is the coordination of large constellations of Earth-observing satellites. Satellite operations have historically followed centralized paradigms, in which a ground controller issues command sequences for each spacecraft well in advance of execution.  Centralized planning is less suitable for emerging time-sensitive use-cases such as disaster response, transient science measurements, and opportunistic observations~\citep{Chien24:Leveraging,Zilberstein25:Real,Chien25:Dynamic}. These use cases, together with rapid improvements in onboard computing, autonomy, and inter-satellite communication~\citep{Rijlaarsdam24:Next,Zilberstein24:Demonstrating}, have motivated recent work on decentralized satellite scheduling~\citep{phillipsauction,picardauction,shreya,zilberstein2025decentralized,Zilberstein26:Large}. 

DCOPs are a successful model for distributed task allocation among spacecraft~\citep{zilberstein2025decentralized}. However, the complete satellite scheduling problem includes complex local constraints such as orbital visibility, sensor slewing, geometric tiling, onboard memory management, and downlink opportunities~\citep{Farges24:Going}. Encoding such structure in a DCOP requires precomputed discretizations and can yield extremely large, dense constraint graphs, especially for modern constellations with hundreds of satellites servicing tens of thousands of observation requests over week-long planning horizons. 

Prior work addressed the scalability challenge through the \textit{neighborhood stochastic search algorithm (NSS)}, a domain-specific geometric decomposition method for the \textit{multi-satellite constellation observation scheduling problem (COSP)}~\citep{zilberstein2025decentralized}. NSS heuristically decomposes the global scheduling problem into smaller subproblems. The decomposition is tightly coupled to the structure of COSP and does not provide a general mechanism for coupling high-level allocation decisions with complex local constraints.
This challenge extends well beyond satellite scheduling. Many large-scale DCOP applications require agents to coordinate a high-level allocation while independently solving rich local planning or scheduling problems.

\paragraph{Our contributions}
We introduce a new decomposition method for large DCOPs, which we refer to as \textit{iterative pricing}.
Our approach uses a high-level meta-DCOP as a discrete allocator.
Given an assignment produced by the DCOP, each agent invokes an independent local solver to determine which assigned tasks can actually be scheduled.
The outcome of the local solver is then fed back into the DCOP through assignment-dependent prices, which discourage future allocations that local agents cannot realize.
Iterative pricing couples global coordination with local reasoning without requiring the local problem to be encoded directly in the DCOP.

Our method is inspired by distributed Lagrangian and pricing methods~\citep{hirayama2009adaptive}, but differs in its decomposition and update structure.
Iterative pricing retains a high-level assignment DCOP, restricts each local solver to the assignments of that DCOP, and updates assignment-specific penalties based on the solution of the local optimization problem.
The local solver can be an exact \textit{mixed integer linear program (MILP)}, a constraint-programming model, a domain-specific heuristic, or a simulator for complex physical constraints.
In the satellite scheduling domain, this allows the global DCOP to reason over high-level request assignment while local solvers handle constraints such as visibility, slewing, memory, downlinking, and geometric tiling.
More generally, iterative pricing provides a generic interface between distributed task allocation and arbitrary local optimization, making it applicable to any DCOP domain in which these two levels are naturally separable.
The idea of two levels of search, a global one for task allocation and local ones for optimizing the handling of tasks, has also been used successfully in other distributed optimization problems beyond DCOPs~\citep{Sandholm93:Implementation}. 

Because iterative pricing repeatedly solves an assignment DCOP, scalable DCOP algorithms become critical.
We revisit the well-known connection between DCOPs and exact potential games~\citep{mgm,chapman2009decentralised,chapman2011benchmarking}, and argue that its implications for modern online learning algorithms as DCOP solvers have received insufficient attention.
We adapt regret matching and several recent accelerations of it such as discounting~\citep{tammelin2014solving,brown2019solving}, optimism~\citep{farina2021faster,syrgkanis2015fast}, and extra-gradient updates for potential games~\citep{anagnostides2025convergence}.
We also analyze how these methods interact with DCOP stabilization techniques such as probability damping and inertia.
While regret matching has recently been applied to DCOPs~\citep{deng2021neural}, that approach relies on pseudo-tree updates that require global coordination.
Our methods instead require only local utility computations and neighbor communication, making them well suited to large-scale decentralized settings.
We discuss further related work on DCOPs, online learning, and satellite scheduling in~\Cref{sec:appendix-related}.

We evaluate our methods on standard graph coloring benchmarks and on large-scale decentralized satellite scheduling problem instances.
On graph coloring, our online learning methods are competitive with representative incomplete DCOP solvers while maintaining low communication overhead.
For the satellite scheduling use case, our iterative pricing substantially improves request completion.
Combining our online learning methods with our iterative pricing framework, we obtain near-optimal performance on real-world decentralized satellite scheduling problem instances, fulfilling over \(99\%\) of observation requests compared with \(87\%\) for state-of-the-art baselines.
Our results establish online learning as a competitive paradigm for incomplete DCOP solving, particularly when paired with iterative pricing.
Our methods also directly apply to the upcoming NASA FAME mission~\citep{fame-spaceops-2025}, which will be the largest demonstration of multiagent AI in space to date. Our experiments are modeled after the operational scenarios that will be tested in flight in 2027, and we seek to deploy our algorithms as part of the FAME mission. 

\section{Preliminaries}

A DCOP is formally defined as a five-tuple $\langle \mathcal{A}, \mathcal{X}, \mathcal{D}, \mathcal{F}, \omega \rangle$ where 
$\mathcal{A}$ is the set of agents,
$\mathcal{X}$ is the set of variables,
and $\mathcal{D}$ is the set of finite domains for the variables, in which each variable $x_i \in \mathcal{X}$ has the corresponding domain $\mathcal{D}_i$.
Each constraint $f\in\mathcal{F}$ has a scope, $\scope(f)\subseteq \mathcal{X}$, and maps assignments of the variables in its scope to a real-valued utility,  $f\colon\prod\limits_{x_i\in\scope(f)} \mathcal{D}_i \rightarrow \mathbb{R}$. The function $\omega\colon\mathcal{X}\rightarrow \mathcal{A}$ assigns each variable to the agent that controls it.

The objective is to find a joint variable assignment maximizing the sum of the utilities
\begin{equation*}
    X^* = \argmax_{X\in \prod_i \mathcal D_i} \sum_{f \in \mathcal{F}} f\left(X|_{\scope(f)}\right),
\end{equation*} 
where $X$ ranges over the possible joint assignment of all variables, and $X|_{\scope(f)}$ denotes the value of $X$ on the variables in $\scope(f)$.

In the most basic DCOP setting, the mapping function $\omega$ is bijective; each agent controls exactly one variable.
This is the case in some DCOP problems, such as graph coloring where each agent controls one node, and most DCOP algorithms are analyzed under this simplification.
However, in many applications, including distributed scheduling, an agent controls many variables and $\omega$ is many-to-one.

A DCOP can be mapped to a \textit{potential game} where each agent $a\in \mathcal{A}$ noncooperatively chooses an assignment $x_a \in \mathcal{D}_a= \prod\limits_{i\in \omega^{-1}(a)}\mathcal{D}_i$ to maximize its local utility $u_a(x_a, X_{-a})$, where $X_{-a}$ denotes the assignments of all other agents~\citep{mgm}.
The connection between DCOPs and exact potential games provides the foundation for the online learning methods developed in this paper.
We define the global potential function $\Phi(X)$ as the sum of all utility functions,
\begin{equation*}
\Phi(X) = \sum_{f \in \mathcal{F}} f(X).
\end{equation*}

A game is an \textit{exact potential game} if for any deviation of an agent from $x_a$ to $x_a'$, the change in an agent $a$'s local utility reflects the change in the global potential:
\begin{equation*}
u_a(x'_a, X_{-a}) - u_a(x_a, X_{-a}) = \Phi(x'_a, X_{-a}) - \Phi(x_a, X_{-a}).
\end{equation*}

In DCOP settings, this property is satisfied when $u_a$ is defined as the sum of utility functions $f\in \mathcal{F}$ that rely on the variable $x_a$.
Every pure Nash equilibrium is a coordinate-wise local optimum of the DCOP, and every global optimum is a pure Nash equilibrium. A pure Nash equilibrium need not be globally optimal. 





\section{Online learning algorithms for DCOPs}

The connection between equilibria of a potential game and local optima of a DCOP motivates using equilibrium finding to solve DCOPs. Online learning is the predominant paradigm used for equilibrium finding in games. 

\paragraph{Regret in DCOPs}
Since our algorithms are based on online learning, we briefly review the notion of \textit{regret} before introducing the algorithms. In an iterative DCOP setting, let $x_a\upt \in \mathcal{D}_a$ be the value assigned to variable $x_a$ by agent $a$ at time step $t$. The local utility $u_a\left(x_a\upt, X_{-a}\upt\right)$ is the sum of all constraints $f \in \mathcal{F}$ involving $x_a$ given the other assignments $X_{-a}\upt$ at that time.

The external regret of agent $a$ for not having chosen a fixed assignment $x'_a \in \mathcal{D}_a$ after $T$ iterations is
\begin{equation*}
R_a^T(x'_a) = \sum_{t=1}^{T} u_a\left(x'_a, X_{-a}\upt \right) - \sum_{t=1}^{T} u_a \left(x_a\upt, X_{-a}\upt \right).
\end{equation*}

An algorithm satisfies the \textit{no-regret property} if, for every agent $a$, the average regret vanishes as $T$ grows,
\begin{equation*}
\lim_{T \to \infty} \frac{1}{T} \max_{x'_a \in \mathcal{D}_a} R_a^T(x'_a) \leq 0.
\end{equation*}
Such algorithms are called \textit{no-regret algorithms}.

\paragraph{Using equilibrium finding for local search in DCOPs}
A key property of exact potential games is that any strategy profile $X$ that is a pure Nash equilibrium is also a local optimum of the potential function $\Phi(X)$.
This observation motivates using equilibrium finding as a \textit{decentralized search procedure}. 
When every agent employs a no-regret algorithm, the empirical distribution of joint assignments converges to the set of coarse correlated equilibria.
However, the no-regret learning guarantees concern empirical regret and equilibrium distributions, and do not guarantee optimality of the individual assignments produced.
Our DCOP algorithms differ from equilibrium finding algorithms in two major ways.
Since we desire variable assignments that produce high-utility solutions rather than a distribution over variable assignments, we sample from the distribution to obtain a solution.
We also do not require that the final solution be sampled from an equilibrium: 
our use of equilibrium finding can be viewed as a heuristic for searching the space of solutions.
We ultimately use the best solution found \textit{during the path to an equilibrium}.
Experimentally, as we will show, the equilibria our algorithms converge to result in high-quality solutions.
In the remainder of this section, we present our DCOP algorithms adapted from the online learning algorithm \textit{regret matching (RM)}~\citep{rm}.

\paragraph{Regret matching (RM)} RM is a no-regret online learning algorithm commonly used for equilibrium finding in games.
While regret-based local-search algorithms have previously been studied for DCOPs~\citep{chapman2009decentralised,chapman2011benchmarking}, more recently, the online learning literature has introduced several RM variants with substantially improved practical performance in large games.
We investigate whether these advances translate into improved decentralized DCOP algorithms.

RM guarantees $O(1/\sqrt{T})$ average external regret after $T$ iterations for settings with bounded utilities and finite actions.
In potential games, recent work establishes additional convergence guarantees toward approximate Nash equilibria under suitable conditions~\citep{anagnostides2025convergence}.
We begin with the standard RM algorithm and then describe the variants evaluated in this paper. 

For simplicity, we present the algorithms in the standard one-variable-per-agent setting, where agent $a$'s action is a value $x_a\in\mathcal{D}_a$.
For multi-variable agents, the same update applies by replacing $x_a$ with a joint local assignment over all variables. 

For any alternative assignment $x\in\mathcal{D}_a$, define the instantaneous regret at time $t$ as $\Delta R_a\upt(x) = u_a(x, X_{-a}\upt) - u_a(x_a\upt, X_{-a}\upt)$. 
The cumulative regret after $t$ iterations is $R_a^{(t)}(x) = R_a^{(t-1)}(x) + \Delta R_a\upt(x)$.
An agent $a$ employing RM computes a distribution, denoted $\sigma_a^{(t)}$ over $\mathcal{D}_a$, proportional to the positive regrets of each value:
\begin{equation*}
\sigma_a\upt(x_a) =
\begin{cases}
\frac{R_a\upt(x_a)^+}{\sum_{x_a' \in \mathcal{D}_a} R_a\upt(x'_a)^+} & \text{if } \sum R_a\upt(x'_a)^+ > 0 \\
\frac{1}{|\mathcal{D}_a|} & \text{otherwise}
\end{cases}
\end{equation*}
where $R_a\upt(x)^+ = \max\left(0, R_a\upt(x)\right)$.
\Cref{alg:regret_matching} presents the full RM algorithm for DCOP.

Like representative incomplete DCOP algorithms, RM requires only local communication with neighboring agents and has a message complexity of $O(|N(a)|)$ per iteration where $N(a)$ is the neighborhood of agent $a$. The computational cost is linear in both the neighborhood size and the local action space. We next describe several RM variants that have accelerated equilibrium finding in large games, and we adapt them to the DCOP setting.

\paragraph{Regret matching+} RM+~\citep{tammelin2014solving} floors cumulative regrets at zero after each update,
$$ R_a\upt(x) \gets \left[R_a^{(t-1)}(x) + \Delta R_a\upt(x)  \right]^+.$$
This prevents actions from accumulating large negative regrets, allowing actions that previously performed poorly to re-enter the strategy more quickly if their counterfactual utilities improve. 

\paragraph{Discounted RM}
\textit{Discounted RM (DRM)}~\citep{brown2019solving} reduces the influence of older regret updates using discount parameters $\alpha$ and $\beta$ for positive and negative cumulative regrets.
Let $d(t,\gamma) = \frac{t^\gamma}{t^\gamma+1}$. DRM updates the cumulative regret by $R_a^{(t-1)}(x) \cdot d(t, \alpha) + \Delta R_a\upt(x)  R_a^{(t-1)}(x)$ when $R_a^{(t-1)}(x) \ge 0$ and $R_a^{(t-1)}(x) \cdot d(t, \beta) + \Delta R_a\upt(x)$ when $R_a^{(t-1)}(x) < 0$.  


\paragraph{Predictive RM} 
\textit{Predictive RM (PRM)}~\citep{farina2021faster} uses a prediction of the next iteration's regret to take a larger step at each strategy update. After updating $R_a^{(t)}$, PRM defines
$$\hat{R}\upt_a(x) \gets R\upt_a(x) + (R\upt_a(x) - R^{(t-1)}_a(x)),$$
and computes $\sigma_a^{(t+1)}$ by applying the RM update with $\widehat{R}_a^{(t)}$.

\paragraph{Increasing-regret PRM}
We also evaluate \textit{increasing-regret PRM (IR-PRM)}~\citep{anagnostides2025convergence}, inspired by recent extra-gradient regret-matching methods for potential games.
These methods modify the predictive update so that the norm of the regret vector does not decrease, which adapts the step size of the regret dynamics while preserving convergence guarantees.

\paragraph{Context-based RM}
In recent work, RM was applied to DCOPs in a \emph{context-based (CB)} manner, which keeps track of regrets conditional on the variable assignments of an agent's neighbors~\citep{deng2021neural}. 
The original implementation performs context-based updates using sequential sampling and backtracking over a pseudo-tree. 
Our implementation instead performs the updates using only local utility computations and communication between neighboring agents, eliminating the pseudo-tree dependency.

The above RM variants can be combined with each other. 
In particular, RM+ combines well with the other variants, and we include hybrids in our experiments. 
For example, the RM+ update with PRM results in the PRM+ algorithm~\citep{farina2021faster}.

\paragraph{FTRL} 
We also compare against \textit{follow the regularized leader (FTRL)} with the \textit{multiplicative weights update (MWU)}, another widely used no-regret online learning algorithm that is outside the RM family.
More details on this algorithm are provided in~\Cref{sec:appendix-ftrl}.

The theoretical guarantees of no-regret algorithms concern regret and equilibrium notions over the sequence of play rather than the quality of individual assignments.
In our implementations, each iteration produces a concrete assignment by sampling from the current strategies, and we evaluate the quality of these assignments empirically.
This allows no-regret dynamics to serve as lightweight decentralized search procedures for incomplete DCOP solving.

\paragraph{DCOP stabilization} 
We additionally evaluate two stabilization heuristics commonly used in incomplete DCOP algorithms: \textit{damping} and \textit{inertia}.
Interestingly, unlike in other DCOP algorithms, we find that these heuristics \textit{decrease} the quality of the solution when combined with online learning algorithms.
We provide more details on these techniques, and prove that RM with inertia violates the no-regret property, in~\Cref{sec:appendix-inertia}.
These results may suggest that moving toward an equilibrium is in fact a strong search heuristic for DCOPs.

\section{Constellation observation scheduling}

We conduct experiments on the canonical DCOP problem of \textit{distributed graph coloring} as well as the real-world \textit{multi-satellite constellation observation scheduling problem (COSP)}, a DCOP formulation of decentralized satellite scheduling introduced by~\citet{zilberstein2025decentralized}.
A COSP instance consists of a scheduling horizon $H=[h_s,h_e]$, a set of satellites $\mathcal{A}$, and a set of observation requests $R$.
Each request $r\in R$ specifies a ground target and a time window $h(r)\subseteq H$ during which the target should be observed. 

For each satellite $a\in \mathcal{A}$, there is a set of candidate observation tasks $S_a$ determined by their orbital mechanics and sensor slewing capabilities.
Each task $s$ is specified by a request $r(s)\in R$, an execution interval $h(s)\subseteq h(r(s))$, and an onboard data volume $m(s)\in\mathbb{R}_{\ge 0}$.
We use a binary decision variable $x_{a,s}\in\{0,1\}$, where $x_{a,s}=1$ indicates that satellite $a$  schedules task $s$.
Each satellite also has a set of mandatory downlink opportunities $\mathcal{L}_a$. A downlink $\ell\in\mathcal{L}_a$ has a contact interval $h(\ell)\subseteq H$ and a maximum data volume $m(\ell)$.

The constraints of the problem dictate the feasibility of scheduling tasks.
Each satellite has a set $\mathcal{C}_a$ of constraints capturing slewing maneuvers, onboard memory management, downlink capacity, and the requirement that a satellite cannot execute overlapping tasks.
For example, if two candidate tasks overlap in time, then they cannot both be scheduled, corresponding to the constraints
$x_{a,s}+x_{a,s'} \le 1$ for all $s,s'\in S_a$ such that $h(s)\cap h(s')\neq\emptyset$.
Let $\mathcal{S}_a^\ell\subseteq\mathcal{S}_a$ denote the tasks whose next downlink opportunity is $\ell$, then memory and downlink capacity impose the constraint $\sum_{s\in S_a^\ell} m(s)x_{a,s} \le \min\{M_a,m(\ell)\}$ where $M_a$ is the total memory budget of satellite $a$. 

The objective is to maximize the number of completed observation requests.
Let
$S(r) = \{(a,s)\mid a\in \mathcal{A},\ s\in S_a,\ r(s)=r\}$
be the set of candidate tasks that can satisfy request $r$. 
The COSP objective is 
$$\mathcal{F}(X)
    =
    \sum_{r\in R}
    \left[
        1-
        \prod_{(a,s)\in S(r)}
        (1-x_{a,s})
    \right],$$
which counts each request at most once, even if multiple satellites could observe it. 
The full scheduling problem can be written as

\[
    X^*
    \in
    \argmax_{X} \mathcal{F}(X) \quad \text{s.t.} \quad X \text{ satisfies } \mathcal{C}_a \quad \forall a.
\]
This formulation induces both a large number of decision variables and a dense constraint graph.
Each satellite can perform many tasks, and local constraints couple tasks within a satellite, while the objective couples satellites that can observe the same target.
In large constellations, many satellites have visibility windows for many of the same requests, producing a dense constraint graph.
The next section focuses on decomposition methods for COSP to make decentralized scheduling tractable.
\section{DCOP decomposition methods}

The direct COSP formulation is too large to solve as a monolithic DCOP, even for moderate-sized constellations.
We therefore develop decomposition techniques motivated by COSP that apply to DCOPs that have a natural decomposition to two levels of optimization.
This approach is similar in spirit to the hierarchical DCOP model MVA~\citep{Fioretto16:Multi} for handling multivariable agents.

We decompose COSP into two interacting problems: a high-level request-assignment DCOP and independent local scheduling problems. 
The top-level DCOP decides which satellite should be responsible for each request.
For request $r\in R$ and satellite $a\in \mathcal{A}$, let $z_{r,a}\in\{0,1\}$ indicate whether $r$ is assigned to $a$.
The assignment DCOP enforces that each request is assigned to at most one satellite\footnote{The problem has the same optimal solutions without this constraint, but our experiments showed that including this constraints had no significant effect on performance.}, $\sum_{a\in \mathcal{A}} z_{r,a} \le 1$ for all $r\in R$. Given an assignment $z$, let $B_a(z)=\{r\in R \mid z_{r,a}=1\}$ be the set of requests assigned to satellite $a$. Satellite $a$ then invokes a local scheduler to determine which requests in $B_a(z)$ can be scheduled such that the constraints of $\mathcal{C}_a$ are satisfied. 



We model the local scheduler for satellite $a$ as an oracle, denoted $\mathcal{O}_a$. Given a bundle of requests $B\subseteq R$ and (possibly) weights $w$, the oracle returns a feasible scheduled subset $Y_a\subseteq B$ together with the corresponding realized schedule, $\widehat{y}_a$. The oracle may be implemented as a MILP, a constraint-programming solver, a domain-specific heuristic, or a simulator that returns a feasible schedule under the local constraints. (In our experiments, we solve the local problem as a MILP.) 
This is in contrast to the monolithic DCOP formulation, where the scheduling problem is itself represented as part of the DCOP formulation.

We study two mechanisms for coupling the assignment DCOP with the local scheduling oracles.
The first is a simple constraint generation baseline in which, over iterations of assignment and scheduling, constraints are learned and added to the assignment DCOP.
The second is our iterative pricing method, which updates the assignment utilities using feedback from the local schedulers.

\subsection{Constraint generation baseline}
\label{sec:constraint-generation}

The assignment DCOP ignores the local scheduling constraints by design for scalability. It may assign a satellite a bundle of requests that cannot be jointly scheduled.  A natural baseline is to iteratively add constraints that prevent the assignment DCOP from repeating infeasible bundles.

At iteration $t$, the assignment DCOP returns an assignment $z\upt$. Each satellite receives the bundle $B_a\upt=B_a(z\upt)$ and invokes its local scheduler. If the scheduler certifies that $B_a\upt$ is feasible, then all assigned requests can be executed by $a$. Otherwise, the algorithm adds the constraint $\sum_{r\in B_a\upt}   z_{r,a} \le |B_a\upt|-1$, which prevents the same infeasible bundle again. 

The constraint is sound when the local scheduler can certify that $B_a\upt$ is in fact infeasible, which can be achieved via a complete solver (\textit{e.g.}, a MILP).
If the local scheduler is heuristic, then the constraint generation is also heuristic, and may eliminate a feasible bundle that the scheduler failed to find.
The procedure terminates when no new constraints are added or a maximum number of iterations are reached.
We outline this procedure in~\Cref{alg:constraint_generation}.
For clarity, the decomposition procedure is shown as synchronous and the call to \textsc{SolveDCOP} denotes the execution of any distributed DCOP solver.
The remaining steps are performed locally by each agent using its assigned bundle and local scheduler.
The pseudocode is centralized only in notation. 

\subsection{Iterative pricing}
\label{sec:iterative-pricing}

Constraint generation gives the assignment DCOP only binary feedback about whether a bundle is infeasible. Iterative pricing accumulates penalties on individual assignments, providing progressively richer feedback to the assignment DCOP over iterations.
Intuitively, if request $r$ is repeatedly assigned to satellite $a$ but cannot be scheduled, then the \textit{price} of assigning $r$ to $a$ increases, making that assignment less attractive in future iterations.

Let $U_r$ be the baseline utility of completing request $r$ ($U_r=1$ in COSP). For each $(r,a)$, iterative pricing maintains a nonnegative price \(\lambda_{r,a}\upt\), initialized to zero. At iteration $t$, the assignment DCOP solves
\[
    z\upt
    \in
    \argmax_z
    \sum_{r\in R}\sum_{a\in \mathcal{A}}
    \left(U_r-\lambda_{r,a}\upt\right)z_{r,a}
\]
subject to $\sum_{a\in \mathcal{A}} z_{r,a}\le 1$ $\forall r\in R$.
Assignments that have previously led to local scheduling failures become less rewarding to the global allocator. Given $z\upt$, each satellite $a$ calls its local scheduler on the bundle $B_a\upt = \{r\in R \mid z_{r,a}\upt=1\}$. The local scheduler returns a feasible subset $Y_a\upt \subseteq B_a\upt$, represented by indicators $y_{r,a}\upt$.  

The local scheduler makes use of weights, defined by $U_r+\lambda_{r,a}\upt$, and solves 
\[
    y_a\upt
    \in
    \argmax_{y_a}
    \sum_{r\in R}
    \left(U_r+\lambda_{r,a}\upt\right)y_{r,a}
\]
subject to $y_a\upt\in\mathcal{C}_a$ and $y_{r,a}\upt\le z_{r,a}\upt$ for all $r\in R$.
The constraint $y_{r,a}\upt\le z_{r,a}\upt$ restricts the scheduler to requests assigned to $a$. The utility $U_r$ encourages the scheduler to complete requests, while the price prioritizes requests that have previously been difficult to reconcile with the global assignment.

After all schedulers return feasible schedules, we compare the assignment $z\upt$ to the schedule $y\upt$. We define $g_{r,a}\upt = z_{r,a}\upt-y_{r,a}\upt$. Since $y_{r,a}\upt\le z_{r,a}\upt$, $g_{r,a}\upt$ is always nonnegative. 
The price update is
\[
    \lambda_{r,a}^{(t+1)}
    =
    \lambda_{r,a}\upt+\alpha \cdot g_{r,a}\upt,
\]
where $\alpha>0$ is a step size. If request $r$ is assigned to satellite $a$ but not scheduled, the price of that assignment increases. If the request is assigned and successfully scheduled, the price remains unchanged. The loop terminates when $z\upt=y\upt$, meaning every assigned request is successfully scheduled, or when a maximum number of iterations is reached. We present this procedure in~\Cref{alg:iterative_pricing}.

\begin{algorithm}[t!]
\caption{Iterative pricing}
\label{alg:iterative_pricing}
\begin{algorithmic}[1]
\STATE \textbf{Input:} DCOP solver \(\textsc{SolveDCOP}\), local schedulers \(\{\mathcal{O}_a\}_{a\in \mathcal{A}}\), utilities \(U\), step size \(\alpha\), iteration limit \(K\)
\STATE Initialize prices \(\lambda_{r,a}^{(1)}\gets 0\) for all \(r\in R,a\in \mathcal{A}\)
\FOR{\(t=1,\dots,K\)}
    \STATE \(z\upt \gets \textsc{SolveDCOP}\big(\{U_r-\lambda_{r,a}\upt\}_{r,a}\big)\)
    \FORALL{\(a\in \mathcal{A}\)}
        \STATE \(B_a\upt \gets \{r\in R \mid z_{r,a}\upt=1\}\)
        \STATE \(w_{r,a}\upt \gets U_r+\lambda_{r,a}\upt\) for all \(r\in B_a\upt\)
        \STATE \( \left( Y_a \upt, \widehat{y}_a \right) \gets \mathcal{O}_a(B_a\upt,w_a\upt) \)
        \STATE Set \(y_{r,a}\upt\gets 1\) if \(r\in Y_a\upt\), and \(0\) otherwise
    \ENDFOR
    \STATE \(\widehat{y}\upt \gets \left(\widehat{y}_a\upt\right)_{a\in\mathcal{A}}\)
    \IF{\(z_{r,a}\upt=y_{r,a}\upt\) for all \(r\in R,a\in \mathcal{A}\)}
        \RETURN{\(\widehat{y}\upt \)}
    \ENDIF
    \FORALL{\((r,a)\in R\times \mathcal{A}\)}
        \STATE \(g_{r,a}\upt \gets z_{r,a}\upt-y_{r,a}\upt\)
        \STATE \(\lambda_{r,a}^{(t+1)}\gets \lambda_{r,a}\upt+\alpha \cdot g_{r,a}\upt\)
    \ENDFOR
\ENDFOR
\RETURN{\(\widehat{y}^{(K)}\)}
\end{algorithmic}
\end{algorithm}

Our method is inspired by distributed Lagrangian pricing methods, which update multipliers associated with violated constraints. Unlike these dual-decomposition methods, iterative pricing retains a high-level assignment DCOP, restricts each local optimizer to the requests selected by that DCOP, and uses assignment-specific penalties. The resulting method is a coordination heuristic rather than an optimization algorithm for a Lagrangian dual. To our knowledge, this is the first price-guided interface between a high-level DCOP allocator and arbitrary local scheduling oracles.
\section{Results}
We begin by comparing the RM and FTRL algorithms against a suite of prominent incomplete DCOP solvers.
We evaluate \textit{DSA-C}, \textit{MGM2}, \textit{GDBA}, and \textit{Maxsum-ADVP}. We then demonstrate the performance of our decomposition frameworks on COSP instances. We compare both the constraint generation baseline and our iterative pricing method to NSS, and evaluate each framework with a suite of DCOP algorithms including the RM variants. We provide details of the hyperparameter selection and compute environment in Sections \ref{sec:appendix-ablations} and \ref{sec:appendix-compute} respectively.


\subsection{Graph coloring}

Graph coloring is a common benchmark for distributed constraint optimization.
We evaluate all algorithms against graph coloring instances on two network topologies: random graphs and scalefree networks.
Random graphs independently construct an edge between each node in the graph with probability $p$.
Scalefree networks model real-world graph structures by iteratively adding vertices to $m$ existing vertices with probability proportional to their degree.
We generate a test set for each graph structure with $n\in\{10,20,30,50,100\}$ number of nodes for a $3$-coloring problem. For each value of $n$, we generate 6 random (two for each choice of $p\in \{\frac4n,\frac5n,\frac6n\}$) and 6 scalefree networks (two for each of $m\in \{2,3,4\}$), for a total of 60 sample problems.
We run each algorithm for 10 trials on all instances.
Each algorithm runs until convergence or until 30 seconds have elapsed.
We tune hyperparameters for each algorithm by conducting a grid search over relevant values.
We provide more details in~\Cref{sec:context-based-comparison}, and a list of hyperparameters used in~\Cref{tab:hyperparam-list-gc}.



\renewcommand{\imgheight}{100 pt}

\ifthenelse{\boolean{isSingleColumn}}{%
        \renewcommand{\imgwidth}{0.5\linewidth}
    }{%
        \renewcommand{\imgwidth}{0.8\linewidth}
    }%
\begin{figure}[t!]
  \centering
    \includegraphics[width=\imgwidth]{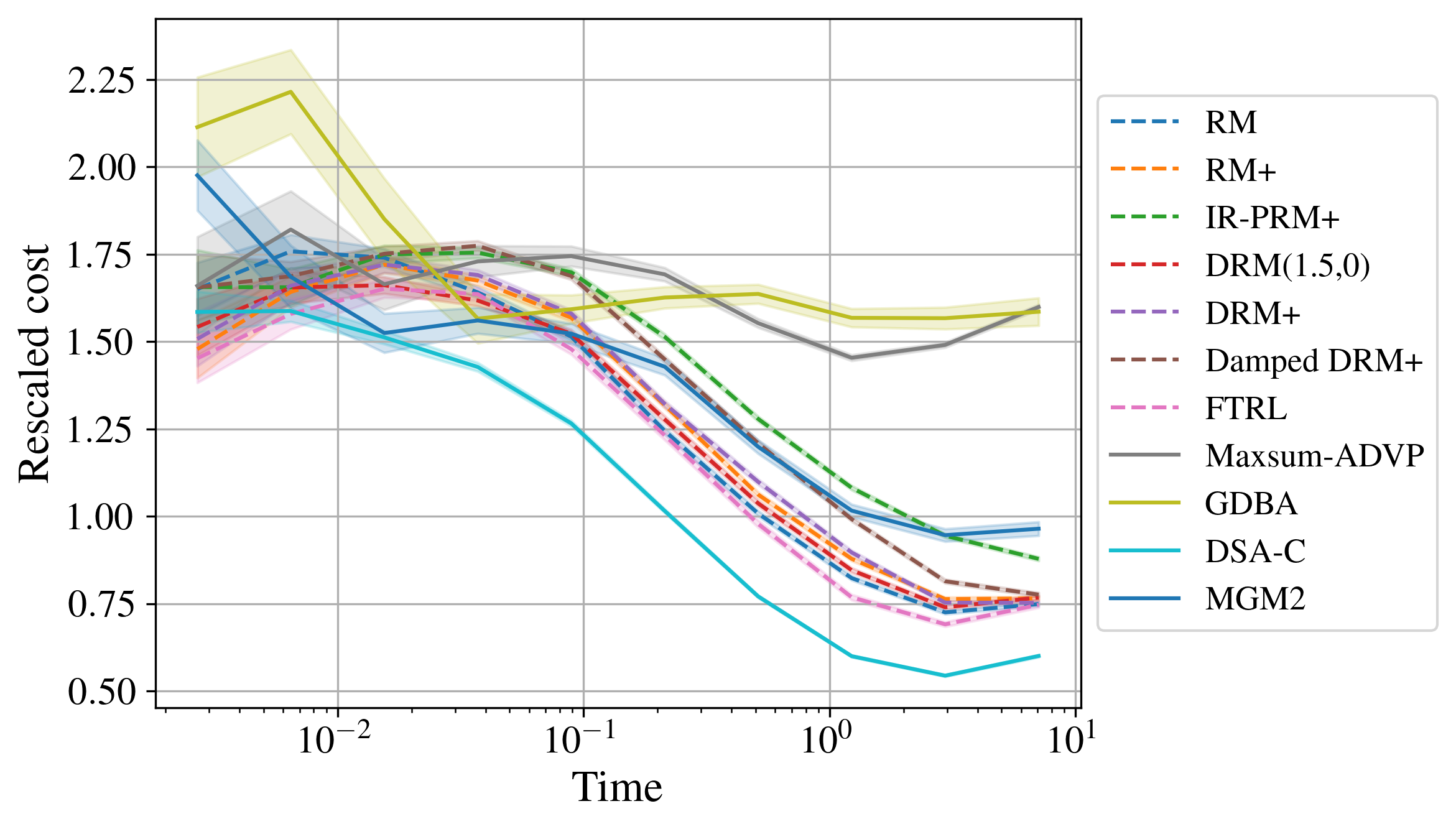}
    \caption{Cost of solution found over time for graph coloring problems across all network types.}
    \label{fig:cost-over-time}
\end{figure}

We report the costs of mid-run solutions found by each algorithm throughout the experiment, and plot these values across time (\Cref{fig:cost-over-time}).
Since the network size $n$ vastly changes the scale of the solution costs, we first rescale the costs so that the experiments with a particular $n$ value have a mean value of 1. 
We use gaussian kernel smoothing to plot each curve, and display the standard error as a shaded region around each curve. 
We show a breakdown of the performance by network type in~\Cref{fig:cost-over-time-appendix}.
We also report the performance by network size in~\Cref{fig:cost-over-time-for-ns}.

We found that across network types and sizes, online learning algorithms performed as well or better than existing incomplete DCOP algorithms, barring DSA.
Most online learning variants performed comparably, including RM+, Damped DRM+, IR-PRM+, and FTRL.
The damped variants of the algorithms underperformed compared to their standard counterparts, indicating that damping is not an effective technique with the RM variants.
Overall, these results support that online learning algorithms are viable for DCOPs.



\subsection{Decentralized satellite scheduling}

We simulate a low-Earth orbit Walker constellation composed of 60 satellites.
The constellation has 8 orbital planes at an $88^\degree$ inclination each containing $6$ satellites. There is an overlay of 2 planes at a $51.6^\degree$ inclination containing 6 satellites each.
This configuration is motivated by the SkySat constellation~\citep{planet}, and is designed to match the number of spacecraft in the NASA FAME mission~\citep{fame-spaceops-2025}.
All satellites are homogeneous; they have a memory capacity of 125 GB and a sensor that can slew to $45^\degree$ off-nadir.
~\Cref{fig:constellation} visualizes the constellation.

\ifthenelse{\boolean{isSingleColumn}}{%
        \renewcommand{\imgwidth}{0.33\linewidth}
    }{%
        \renewcommand{\imgwidth}{0.5\linewidth}
    }%
\begin{figure}[t!]
    \centering
    \includegraphics[width=\imgwidth]{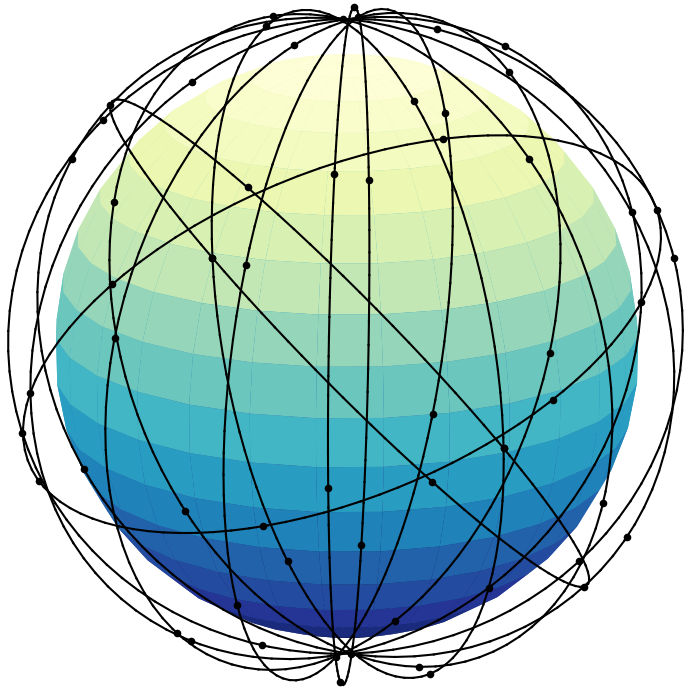}
    \caption{The 60-satellite Walker constellation.}
    \label{fig:constellation}
\end{figure}

We also simulate two downlink stations: the ASF Near Space Network Satellite Tracking Ground
Station and the Guam Remote Ground Terminal System.
A downlink is modeled by a bit stream of 62.5 MB/s.

For each simulation, we randomly select a horizon of six hours within a one week interval. Six hours corresponds to roughly four full orbits of Earth per satellite. We generate a request set composed of repeat observations of a random subset of $634$ of the most populous, globally distributed cities.
For each ground target, we add a request to observe it in the first half, second half, or both halves of the horizon.

\ifthenelse{\boolean{isSingleColumn}}{%
        \renewcommand{\imgwidth}{0.5\linewidth}
    }{%
        \renewcommand{\imgwidth}{0.7\linewidth}
    }%
\begin{figure}[b!]
    \centering
    \includegraphics[width=\imgwidth]{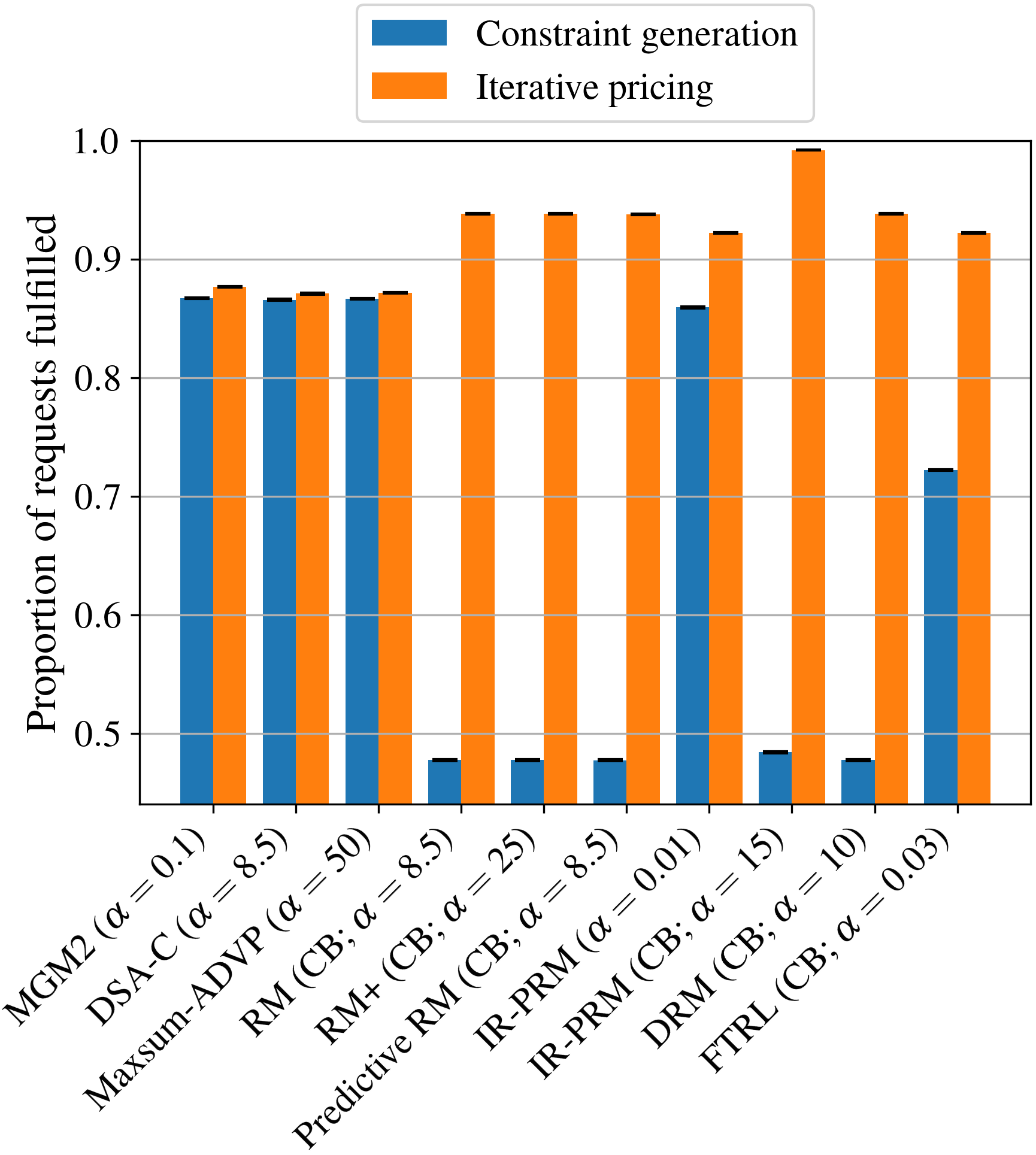}
    \caption{Average quality of satellite scheduling solution found. Standard error of the sample mean is displayed.}
    \label{fig:sat-sched-utility}
\end{figure}

We generate 20 problem instances (observation campaigns) for evaluation, and run each algorithm within each framework for 20 trials on each instance.
We report the utility of the final solution achieved by each algorithm in~\Cref{fig:sat-sched-utility}. 
We use the context-based variants of the RM algorithms, as this resulted in the best performance when paired with the iterative pricing framework.
Context-based variants had the opposite effect under constraint generation, so we display the standard variant of IR-PRM as a representative.
We provide more details on the context-based results in~\Cref{sec:context-based-comparison}. 
We run each algorithm for 25 iterations, which we found to be enough for convergence (\Cref{sec:max-iterations}).
The iterative pricing step size, $\alpha$, is also a tunable parameter, and we show these results in~\Cref{sec:step-size-tuning}. 
With each algorithm, we report the optimal $\alpha$ found in the iterative pricing framework.
We provide the final list of hyperparameters used in~\Cref{tab:hyperparam-list-ss}.






The best algorithm within the iterative pricing framework was context-based IR-PRM, which fulfilled 99.2\% of requests. 
The constraint generation framework with MGM2, fulfilled only 86.7\% of the observation requests.
The iterative pricing framework performed significantly better than constraint generation with all algorithms apart from DSA-C and Maxsum-ADVP.
This difference is likely due to the combinatorial infeasibility of restricting enough schedules to force the algorithm to converge to a better solution.
Within the iterative pricing framework, all online learning algorithms performed well, fulfilling over 90\% of requests.
The other incomplete DCOP algorithms failed to reach 90\% request completion.
The online learning variants performed substantially better within our framework.





Finally, we compare our frameworks to NSS (\Cref{tab:nss-comparison}). 
We report the highest satisfaction variant of each algorithm, and use NSS with the GND(2) heuristic.
Our iterative pricing approach was near-optimal and satisfied over $99\%$ of requests.
NSS and the constraint generation framework only achieved around $87\%$ satisfaction.
However, NSS, which focuses on scalability, reduced the message volume substantially. 
This indicates a natural tradeoff between our method and NSS: for closer-to-optimal solutions, one should select our iterative pricing framework, and for applications in which very low message volume is important, NSS is more appropriate.

\ifthenelse{\boolean{isSingleColumn}}{%
        \renewcommand{\tabwidth}{0.75\linewidth}
    }{%
        \renewcommand{\tabwidth}{\linewidth}
    }%
\begin{table}[!htbp]
    \centering
    \resizebox{\tabwidth}{!}{
    \begin{tabular}{lcc}
        \toprule
        \textbf{Framework} & \textbf{Proportion of requests fulfilled} & \textbf{Message volume}\\
        \midrule
        NSS &  0.870 & \textbf{84,180} \\\hline
        Constraint generation & \multirow{2}*{0.867} & \multirow{2}*{696,087}\\
        (+ MGM2) & &\\
        \hline
        Iterative pricing & \multirow{2}*{\textbf{0.992}} & \multirow{2}*{1,331,750} \\
        (+ CB IR-PRM) &  & \\
        \bottomrule
    \end{tabular}}
    \caption{Average results of highest satisfaction variant of each framework across 20 problem instances.}
    \label{tab:nss-comparison}
\end{table}

\section{Conclusions and future work}

We presented two complementary contributions toward scalable distributed constraint optimization.
We demonstrated that modern online learning algorithms can be adapted to search and are a competitive family of incomplete DCOP solvers.
These methods achieved solution quality comparable to the predominant prior incomplete DCOP algorithms on graph coloring benchmarks. We introduced iterative pricing, a new price-guided decomposition framework for DCOPs.
Rather than requiring complex local planning problems to be encoded directly within the DCOP, iterative pricing couples a high-level assignment DCOP with independent local optimizers through prices. On large-scale decentralized satellite scheduling problems, our framework substantially improved request completion over prior decomposition methods.

Although our experiments focused on satellite scheduling, our framework is applicable whenever a distributed allocation problem naturally decomposes into a global assignment problem together with rich local planning or scheduling problems. We anticipate that this abstraction will be useful across distributed planning domains, including mobile sensor teams~\citep{ca-dcop-mst}, UAV coordination~\citep{uavs-coord}, and trucking task allocation, routing, and scheduling~\citep{Sandholm93:Implementation}.

\section*{Acknowledgments}

Tuomas Sandholm and his PhD students Itai Zilberstein and Pranav Rajbhandari are supported by NIH award A240108S001, the Vannevar Bush Faculty Fellowship ONR N00014-23-1-2876, and National Science Foundation grant RI-2312342. Itai Zilberstein is also supported by the NSF Graduate Research Fellowship Program under grant DGE2140739. Any opinions, findings, and conclusions or recommendations expressed in this material are those of the author(s) and do not necessarily reflect the views of the funding agencies.
\bibliography{references}

\newpage 

\appendix
\setcounter{table}{0}
\renewcommand{\thetable}{A\arabic{table}}

\setcounter{algorithm}{0}
\renewcommand{\thealgorithm}{A\arabic{algorithm}}

\setcounter{figure}{0}
\renewcommand{\thefigure}{A\arabic{figure}}

\section{Additional related work}
\label{sec:appendix-related}

\paragraph{Distributed constraint optimization}
Optimal algorithms for DCOP typically operate on the \emph{pseudo-tree} representation of a DCOP, incurring exponential time and message complexity by performing exhaustive search~\citep{adopt,syncbb,gershman2009asynchronous}.
Incomplete DCOP algorithms are typically based on local search, sampling, or inference. We compare against representative methods from these families, including the \textit{distributed stochastic search algorithm (DSA)}~\citep{dsa}, \textit{generalized distributed breakout algorithm (GDBA)}~\citep{gdba}, \textit{maximum gain messaging (MGM)}~\citep{mgm},  \textit{D-Gibbs}~\citep{dgibs}, and \textit{Maxsum}~\citep{maxsumadvp}.

To handle complex local problem spaces, prior work introduced the \textit{multi-variable agent (MVA)} method for DCOPs~\citep{Fioretto16:Multi}.
MVA groups local variables controlled by the same agent and utilizes a centralized solver within the agent before participating in message passing.
While MVA significantly improves scalability by allowing agents to solve local optimization problems internally, the local variables and constraints must still be explicitly represented within the DCOP model.

Our approach is motivated by the \textit{distributed Lagrangian relaxation protocol (DisLRP)}~\citep{hirayama2009adaptive}, a pricing penalty method originally developed for  the generalized mutual assignment problem.
DisLRP updates Lagrange multipliers associated with relaxed constraints. Our approach is similarly inspired by pricing methods, but differs in three important respects. Iterative pricing retains a high-level assignment DCOP rather than solving a dual decomposition. Second, local schedulers are restricted to the requests assigned by the DCOP. And finally, prices are updated from scheduling outcomes rather than explicit constraint violations in the global model.

Our decomposition is also conceptually related to decomposition techniques from mathematical optimization, particularly \textit{Logic-based Benders decomposition (LBBD)}~\citep{Hooker03:Logic}. LBBD separates a global master problem from richer local optimization subproblems that may be solved by arbitrary optimization procedures. 

The connection between DCOPs and \emph{potential games} was made early on in the literature, and regret-based local-search algorithms have been investigated for DCOPs~\citep{chapman2009decentralised,chapman2011benchmarking}. 
More recently, \citet{deng2021neural} proposed a context-based regret-matching algorithm that performs updates using sequential sampling and backtracking over a pseudo-tree. Our algorithms instead use only local utility evaluations and neighbor communication. 

Recent advances in regret minimization such as optimism~\citep{farina2021faster}, reweighting~\citep{brown2019solving}, and discounting~\citep{tammelin2014solving} have substantially improved practical equilibrium finding in large games. 
Theoretically, optimistic learning rates are known to accelerate convergence in potential games~\citep{syrgkanis2015fast}.
However, the practical efficacy of these predictive dynamics in large multiagent systems with synchronous updates is less understood.
Regret matching has also been recently studied theoretically in the context of potential games, where convergence guarantees were obtained~\citep{anagnostides2025convergence}.

\paragraph{Satellite scheduling} Satellite observation scheduling is primarily modeled as an optimization problem involving computational geometry, constrained task allocation, and coordination among downlink stations, satellites, and operation centers.
The vast majority of prior work has focused on centralized paradigms for scheduling, and this is the standard in deployed approaches~\citep{globus,augenstein2016optimal,nag2018scheduling,he2018improved,shah2019scheduling,aeossp,squillaci2021managing,boerkoel2021efficientsatellite,eddy2021maximum,squillaci2023scheduling,sat-task-plan-large-areas,Barrault25:Hybridizing,Kim26:Free}.
Decentralized approaches include auction-based methods \citep{picardauction,phillipsauction} and heuristic search-based methods \citep{shreya,bonnet2007collaboration,bonnet2008coordination,zilberstein2025decentralized}.
We leverage the DCOP formulation of decentralized satellite scheduling from~\citet{zilberstein2025decentralized}, referred to as the \textit{multi-satellite constellation observation scheduling problem (COSP)}.

COSP is a challenging application of DCOP methods due to its scale and structure.
COSP instances are typically composed of tens to hundreds of satellites and hundreds to thousands of requests, resulting in up to millions of decision variables.
Due to the short orbital period of low-Earth orbiting satellites, each agent gets many visibility windows of ground targets. The resulting constraint graph contains many high-degree nodes and dense subgraphs. These factors make DCOP approaches that rely on agents communicating with neighboring agents in the constraint graph computationally challenging due to the high degrees in the graph.
 
To solve COSP, prior work used decomposition-based methods to decompose the constraint graph by partitioning agents and requests in order to run DSA; this approach is referred to as \textit{neighborhood stochastic search (NSS)}~\citep{zilberstein2025decentralized}. NSS is empirically effective, but its geometric decomposition is tailored to COSP. In contrast, our method retains a global DCOP and uses feedback from local scheduling oracles to update assignment prices. This provides a more general interface between distributed task allocation and local constraint reasoning, and enables local scheduling problems to be modeled outside the DCOP representation.

It is also worth noting that our abstraction of the local scheduling problem is directly compatible with other single-agent technologies for spacecraft planning such as dynamic targeting~\citep{Zilberstein25:Real,Chien25:Dynamic,Kangaslahti26:Dynamic}.

\section{Inertia and damping}
\label{sec:appendix-inertia}

\textit{Damping} smooths changes in an agent's mixed strategy through the update $ \sigma_a^{(t+1)} \leftarrow (1-\lambda) \sigma_a^{(t)} + \lambda\sigma_a^{(t+1)}$ for $\lambda \in [0,1]$.
\textit{Inertia} reduces thrashing by limiting how often agents change their assignments.
For $p_I\in[0,1]$, agent $a$ keeps its previous action with probability $p_I$, $x_a^{(t+1)} \gets x_a\upt$. 
And, with probability $1-p_I$ an agent samples from the current strategy, $x_a^{(t+1)} \sim \sigma_a^{(t+1)}$.

We show that inertia does not generally preserve no-regret guarantees by giving a two-action counterexample in which the learner incurs linear external regret.

Consider a single agent with two actions, denoted \(e_1=(1,0)\) and \(e_2=(0,1)\). 
At each iteration \(t\ge 2\), define the utility vector adversarially as a function of the agent's previous realized action:
\[
u^{(t)} =
\begin{cases}
    (1,-1) & \text{if } x^{(t-1)}=e_2,\\
    (-1,1) & \text{if } x^{(t-1)}=e_1.
\end{cases}
\]
Thus, the action opposite to the agent's previous action receives utility \(1\), while repeating the previous action receives utility \(-1\).

Suppose the non-inertial update would select the opposite of the previous action, but inertia forces the agent to keep its previous action with probability \(p\). 
Then, conditioned on the history before round \(t\), the expected utility of the inertialized action is
\[
    \mathbb{E}\left[u^{(t)}(x^{(t)})\mid \mathcal{H}_{t-1}\right]
    =
    (1-p)\cdot 1 + p\cdot (-1)
    =
    1-2p.
\]
Therefore, for \(T\) rounds,
\[
    \mathbb{E}\left[\sum_{t=2}^{T} u^{(t)}(x^{(t)})\right]
    =
    (T-1)(1-2p).
\]

Now consider the best fixed action in hindsight. 
For every realized utility vector \(u^{(t)}\), the two action utilities sum to zero:
\[
    u^{(t)}(e_1)+u^{(t)}(e_2)=0.
\]
Hence, for every realized history,
\[
    \max_{i\in\{1,2\}} \sum_{t=2}^{T} u^{(t)}(e_i)
    \ge
    \frac{1}{2}\sum_{i=1}^{2}\sum_{t=2}^{T}u^{(t)}(e_i)
    =
    0.
\]
It follows that the expected external regret is at least
\[
    \mathbb{E}\left[
    \max_{i\in\{1,2\}} \sum_{t=2}^{T}u^{(t)}(e_i)
    -
    \sum_{t=2}^{T}u^{(t)}(x^{(t)})
    \right] \ge (2p-1)(T-1).
\]
For any \(p>1/2\), inertia can incur linear external regret.




\section{Follow the regularized leader}
\label{sec:appendix-ftrl}

While RM focuses on the difference between counterfactual and actual utilities, \textit{follow the regularized leader (FTRL)} is a general framework where agents choose a strategy that would have performed best across all previous iterations, modified by a regularization term to maintain stability.
In FTRL, the strategy for agent $a$ at time $t+1$ is defined as:

\begin{equation*}
\sigma_a^{(t+1)} = \smash{\argmax_{\sigma \in \Delta(\mathcal{D}_a)}} \left( \sum_{\tau=1}^t \sum_{x \in \mathcal{D}_a} \sigma(x) u_a(x, X_{-a}^{(\tau)}) - \mathcal{R}(\sigma) \right)
\end{equation*}

where $\mathcal{R}(\sigma)$ is a strongly convex regularizer.

The \textit{multiplicative weights update (MWU)} algorithm is one of the most common instances of FTRL.
It utilizes negative Shannon entropy as the regularizer:

\begin{equation*}
    \mathcal{R}(\sigma) = \frac{1}{\eta} \sum_{x \in\mathcal{D}_a} \sigma(x)\ln \sigma(x)
\end{equation*}
where $\eta > 0$ is the learning rate.
The MWU strategy at iteration $t+1$ is then computed proportional to the softmax: 

\begin{equation*}
\sigma_a^{(t+1)}(x) = \frac{\exp\left(\eta R_a^t(x)\right)}{\sum_{x' \in \mathcal{D}_a} \exp\left(\eta R_a^t(x')\right)}.
\end{equation*}

MWU is also a no-regret online learning algorithm with the same $O(|N(a)|)$ message complexity as RM.
However, MWU will maintain a strictly positive probability for every variable assignment.

\section{Additional pseudocode}

\begin{algorithm}[h!]
\caption{Distributed regret matching (RM) for agent $a$}
\label{alg:regret_matching}
\begin{algorithmic}[1]
\STATE \textbf{Initialize:} 
\STATE Cumulative regrets $R_a(x) \gets 0, \forall x \in \mathcal{D}_a$
\STATE Select initial assignment $x_a^{(1)} \in \mathcal{D}_a$ uniformly at random
\STATE Send $x_a^{(1)}$ to all neighbors $N(a)$
\FOR{each iteration $t = 1, \dots, T$}
    \STATE \textbf{Wait} for messages $x_j^{(t)}$ from all neighbors $j \in N(a)$
    \STATE Form the current neighborhood assignment $X_{-a}^{(t)}$
    \STATE Compute actual utility: $u_a(x_a^{(t)}, X_{-a}^{(t)})$
    \FORALL{$x \in \mathcal{D}_a$}
        \STATE Compute counterfactual utility: $u_a(x, X_{-a}^{(t)})$
        \STATE Update cumulative regret:
        \STATE $R_a(x) \gets R_a(x) + (u_a(x, X_{-a}^{(t)}) - u_a(x_a^{(t)}, X_{-a}^{(t)}))$
    \ENDFOR
    \STATE $S_a \gets \sum_{x' \in \mathcal{D}_a} \max(0, R_a(x'))$
    \FORALL{$x \in \mathcal{D}_a$}
        \IF{$S_a > 0$}
            \STATE $\sigma_a^{(t+1)}(x) \gets \frac{\max(0, R_a(x))}{S_a}$
        \ELSE
            \STATE $\sigma_a^{(t+1)}(x) \gets \frac{1}{|\mathcal{D}_a|}$
        \ENDIF
    \ENDFOR
    \STATE Sample next assignment $x_a^{(t+1)} \sim \sigma_a^{(t+1)}$
    \STATE Send $x_a^{(t+1)}$ to all neighbors $N(a)$
\ENDFOR
\end{algorithmic}
\end{algorithm}

\begin{algorithm}[h!]
\caption{Constraint generation}
\label{alg:constraint_generation}
\begin{algorithmic}[1]
\STATE \textbf{Input:} DCOP solver \(\textsc{SolveDCOP}\), local schedulers \(\{\mathcal{O}_a\}_{a\in \mathcal{A}}\), utilities \(U\), iteration limit \(K\)
\STATE Initialize constraint set \(\mathcal{G}\gets\emptyset\)
\FOR{\(t=1,\dots,K\)}
    \STATE \(z\upt \gets \textsc{SolveDCOP}(\mathcal{G})\)
    \FORALL{\(a\in \mathcal{A}\)}
        \STATE \(B_a\upt \gets \{r\in R \mid z_{r,a}\upt=1\}\)
        \STATE \( \left( Y_a\upt, \widehat{y}\upt_a \right) \gets \mathcal{O}_a(B_a\upt,U)\)
        \IF{\(Y_a\upt \neq B_a\upt \)}
            \STATE Add constraint \(\sum_{r\in B_a\upt} z_{r,a}\le |B_a\upt|-1\) to \(\mathcal{G}\)
        \ENDIF
    \ENDFOR
    \STATE \(\widehat{y}\upt \gets \left(\widehat{y}_a\upt\right)_{a\in\mathcal{A}}\)
    \IF{no constraints were added}
        \RETURN{\(\widehat{y}\upt\)}
    \ENDIF
\ENDFOR
\RETURN{\(\widehat{y}^{(K)}\)}
\end{algorithmic}
\end{algorithm}

\section{Ablation studies}
\label{sec:appendix-ablations}

\subsection{Hyperparameter choice for graph coloring}
\label{sec:appendix-hyperparameter-graph-coloring}

For our initial graph coloring study, we consider a grid of relevant hyperparameter values for each tested algorithm, and select the best performing value to use for our main experiment.
Results for hyperparameter searches are obtained from generated networks with $n\in\{10,20,30,50,100\}$ number of nodes.
For each value of $n$, we generate 5 random and 5 scalefree networks, and run 10 trials on our 50 instances.
We report mid-run solution costs on a log time scale.

\subsubsection{RM Damping}
When comparing different levels of damping in the RM algorithm, we find that using no damping performs the best (\Cref{fig:gc_hp_RM_damp}).

\ifthenelse{\boolean{isSingleColumn}}{%
    \renewcommand{\imgwidth}{0.5\linewidth}
}{%
    \renewcommand{\imgwidth}{\linewidth}
}%
\begin{figure}[h!]
    \centering
    \includegraphics[width=\imgwidth]{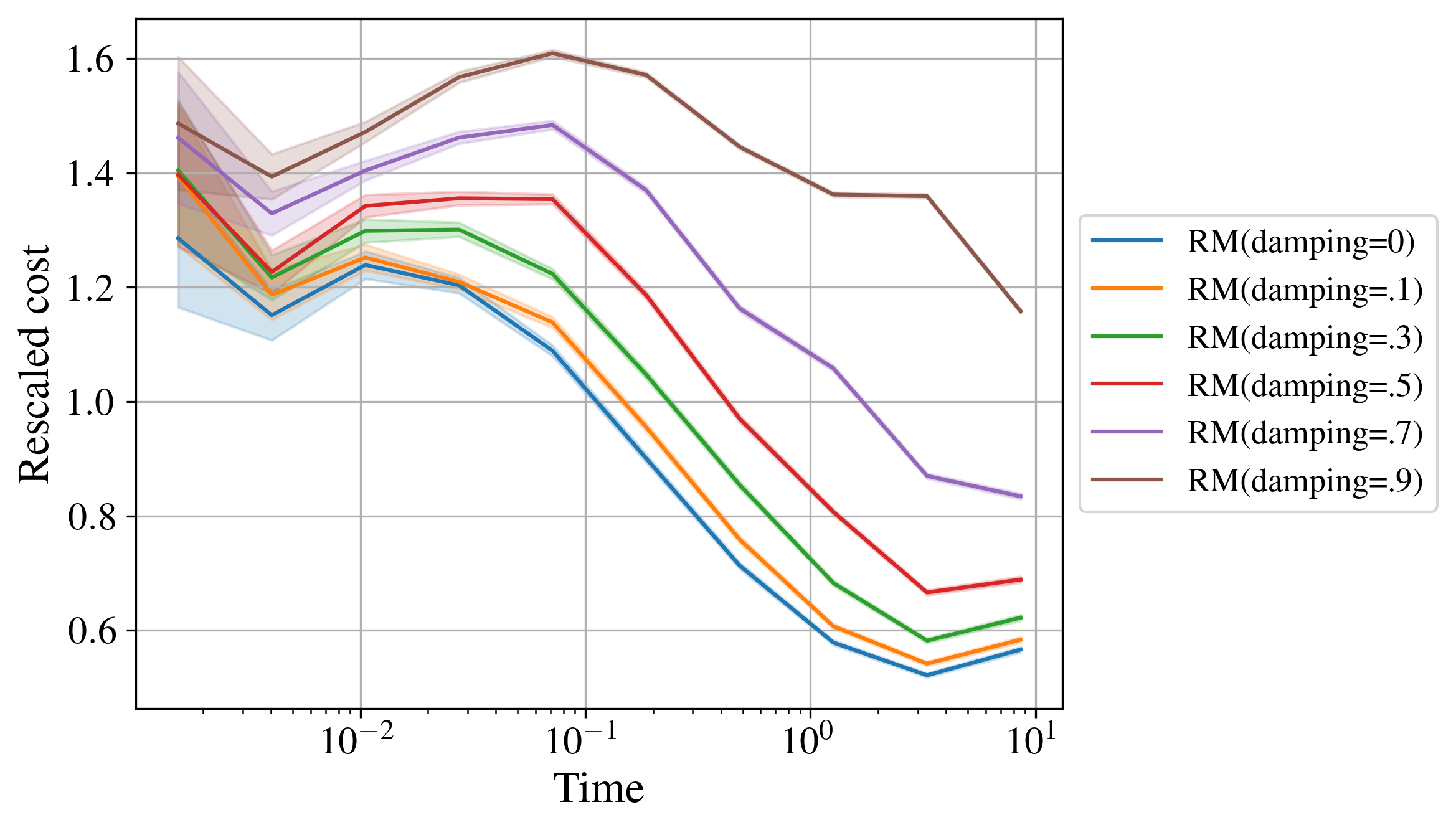}
    \caption{Cost of solution found over time for different levels of damping in RM.}
    \label{fig:gc_hp_RM_damp}
\end{figure}

\subsubsection{RM Inertia}
\begin{figure}[h!]
    \centering
    \includegraphics[width=\imgwidth]{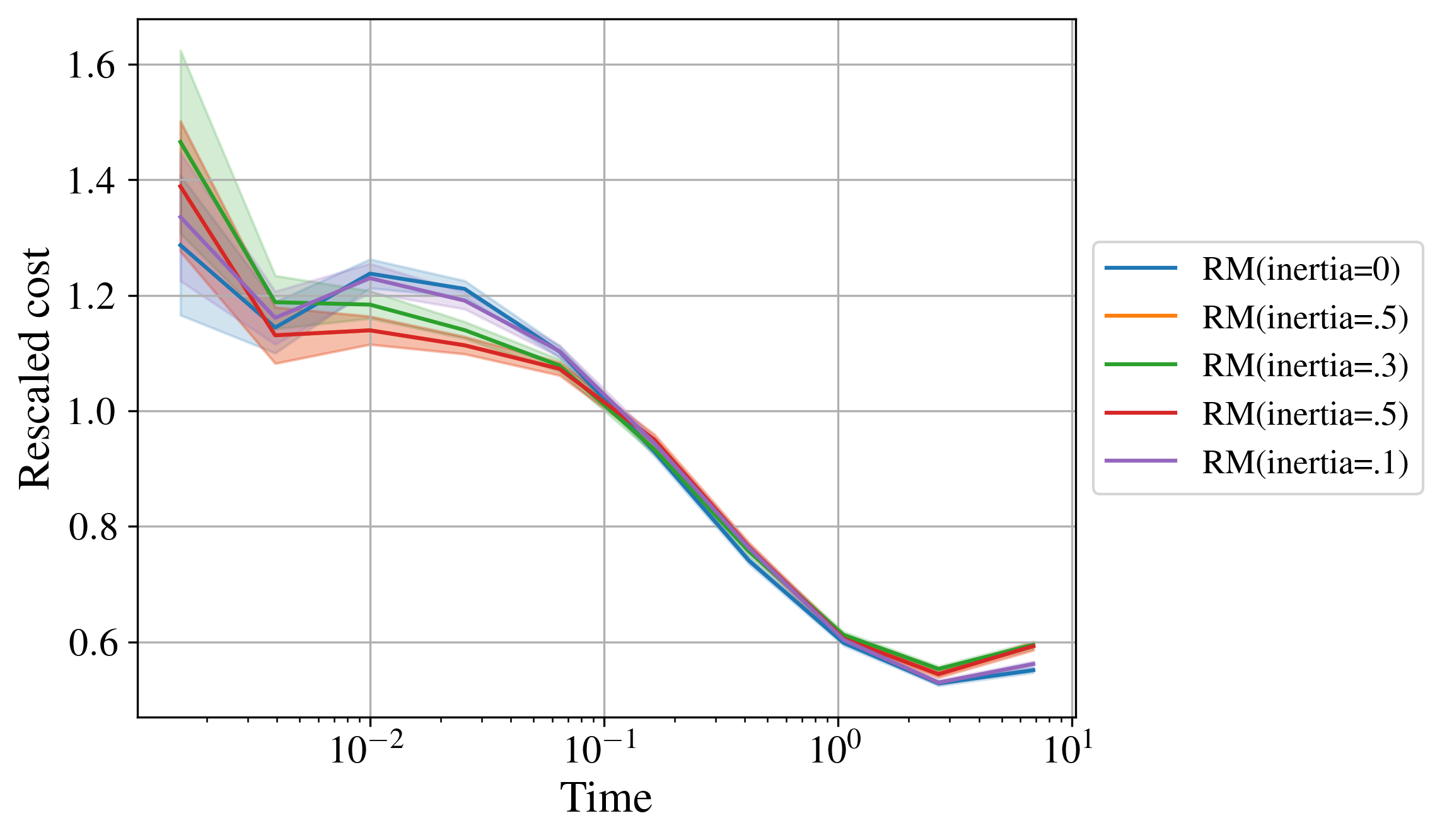}
    \caption{Cost of solution found over time for different levels of inertia in RM.}
    \label{fig:gc_hp_RM_inert}
\end{figure}

When comparing different levels of inertia in the RM algorithm, we find that using no inertia performs the best (\Cref{fig:gc_hp_RM_inert}).
Combined with the previous result, this indicates that when paired with online algorithms, the increased stability of damping and inertia algorithm variants comes with a loss in performance.

\subsubsection{DRM $\alpha$ and $\beta$}
\begin{figure}[h!]
    \centering
    \includegraphics[width=\imgwidth]{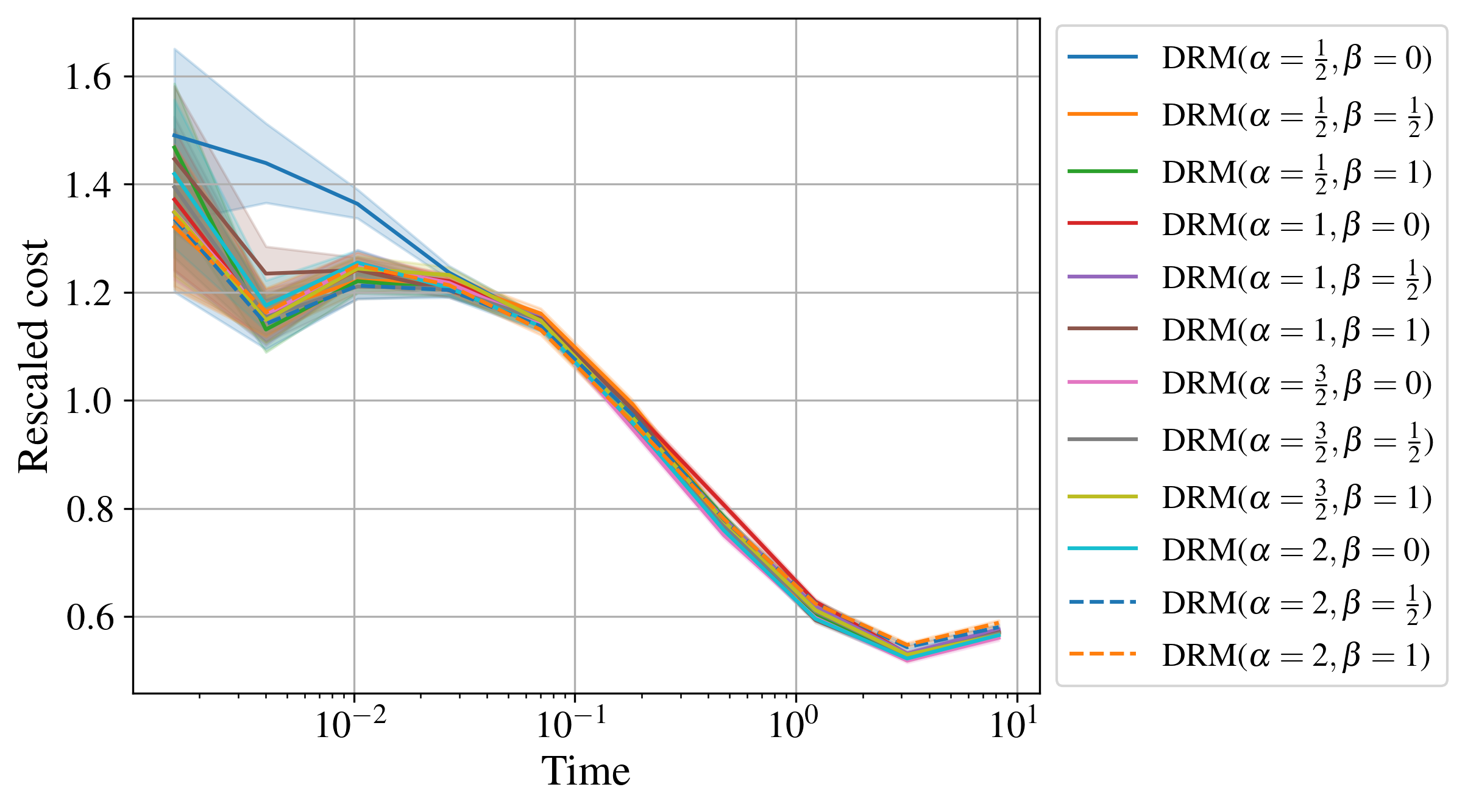}
    \caption{Cost of solution found over time for different $\alpha$ and $\beta$ values in DRM.}
    \label{fig:gc_hp_DRM}
\end{figure}

When comparing different values for $\alpha$ and $\beta$ in the DRM algorithm, we find there is no major difference in performance (\Cref{fig:gc_hp_DRM}).
In our main experiments, we use $\alpha=1.5$, $\beta=0$ since these values are often used.

\subsubsection{DSA}
\begin{figure}[h!]
    \centering
    \includegraphics[width=\imgwidth]{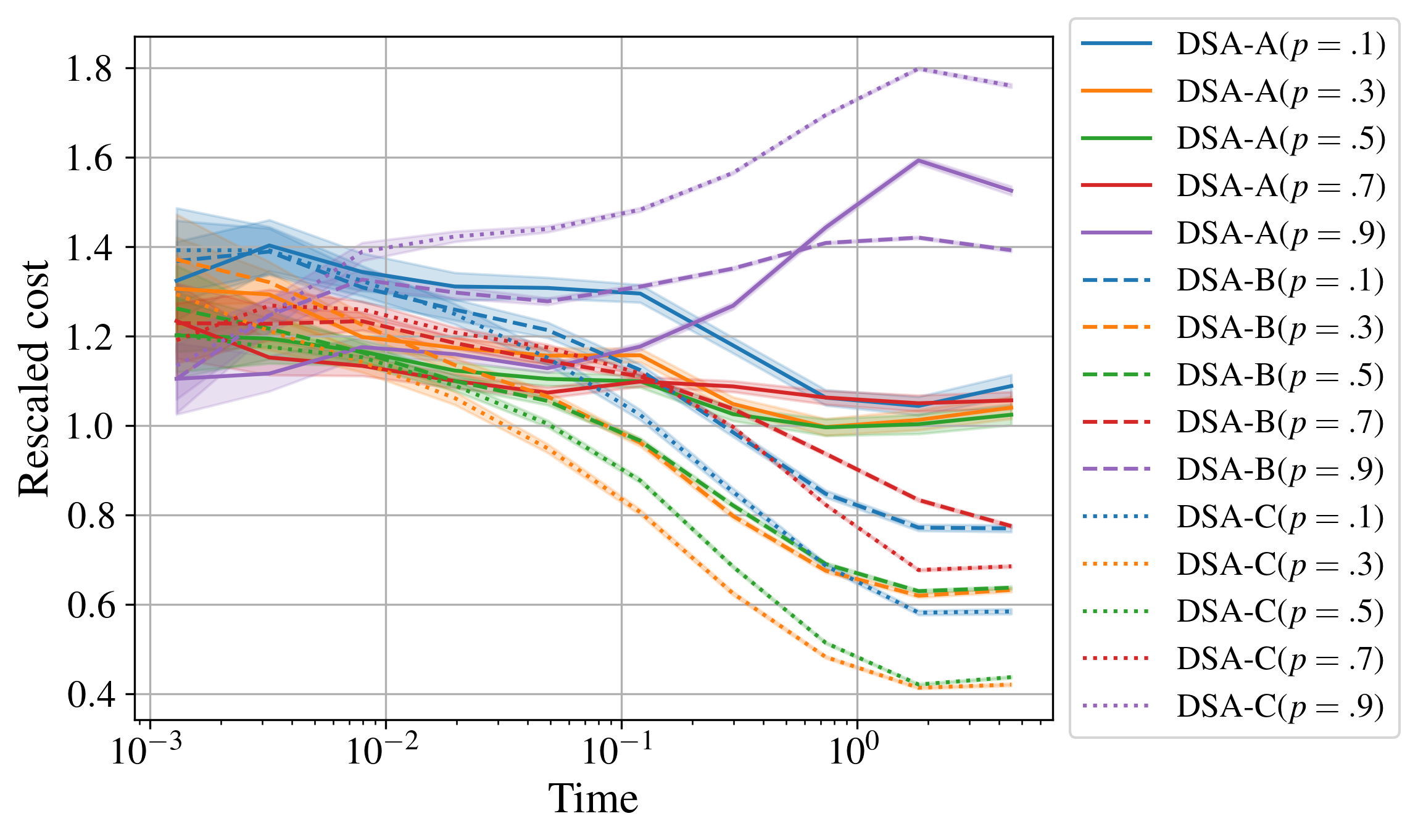}
    \caption{Cost of solution found over time for different $p$ values in DSA, across different variants.}
    \label{fig:gc_hp_DSA}
\end{figure}

When comparing different variants of DSA (as well as different $p$ parameters), we find that DSA-C with $p=0.3$ or $p=0.5$ perform best (\Cref{fig:gc_hp_DSA}).

\subsubsection{FTRL}
\begin{figure}[h!]
    \centering
    \includegraphics[width=\imgwidth]{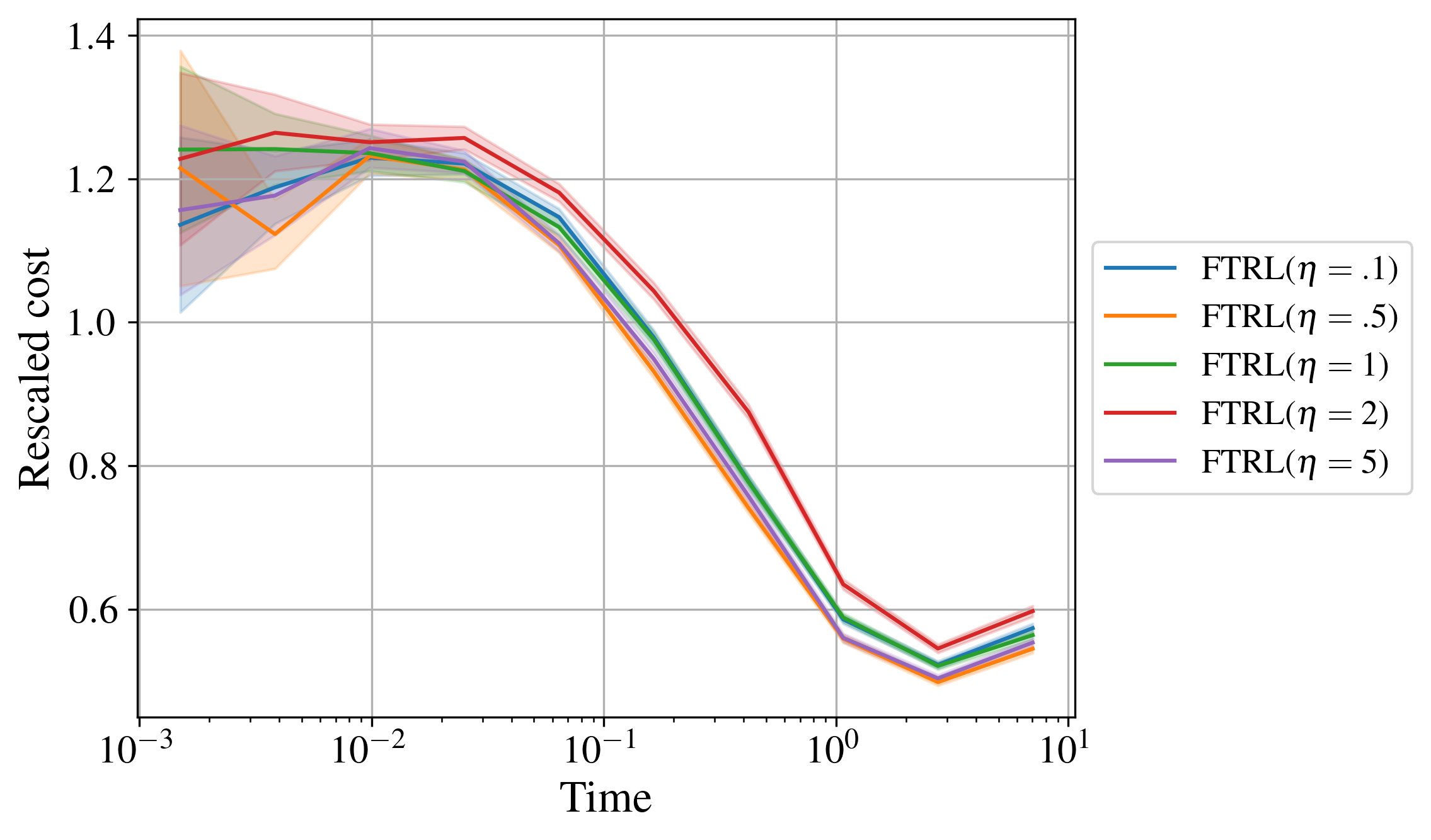}
    \caption{Cost of solution found over time for different $\eta$ values in FTRL.}
    \label{fig:gc_hp_FTRL}
\end{figure}

When comparing different values for $\eta$ in the FTRL algorithm, we find there is no major difference in performance (\Cref{fig:gc_hp_FTRL}).
In our main experiments, we use $\eta=1$ since this value is often used.
\subsubsection{Maxsum Damping}
\begin{figure}[h!]
    \centering
    \includegraphics[width=\imgwidth]{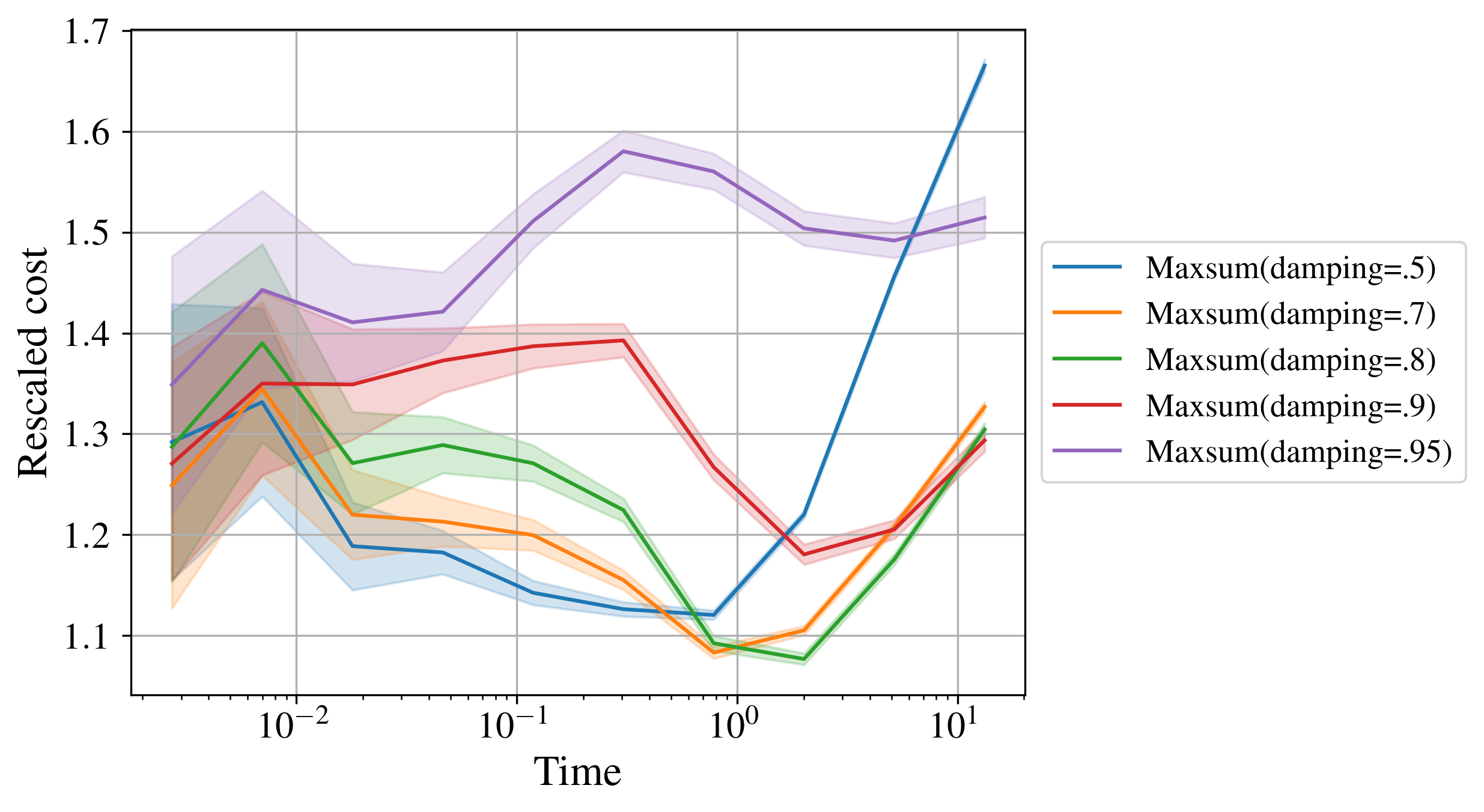}
    \caption{Cost of solution found over time for different levels of damping in Maxsum.}
    \label{fig:gc_hp_MS_damp}
\end{figure}

When comparing different levels of damping in the Maxsum algorithm, we find that using a damping parameter of $0.8$ performs the best (\Cref{fig:gc_hp_MS_damp}).
\subsubsection{Maxsum Stability}
\begin{figure}[h!]
    \centering
    \includegraphics[width=\imgwidth]{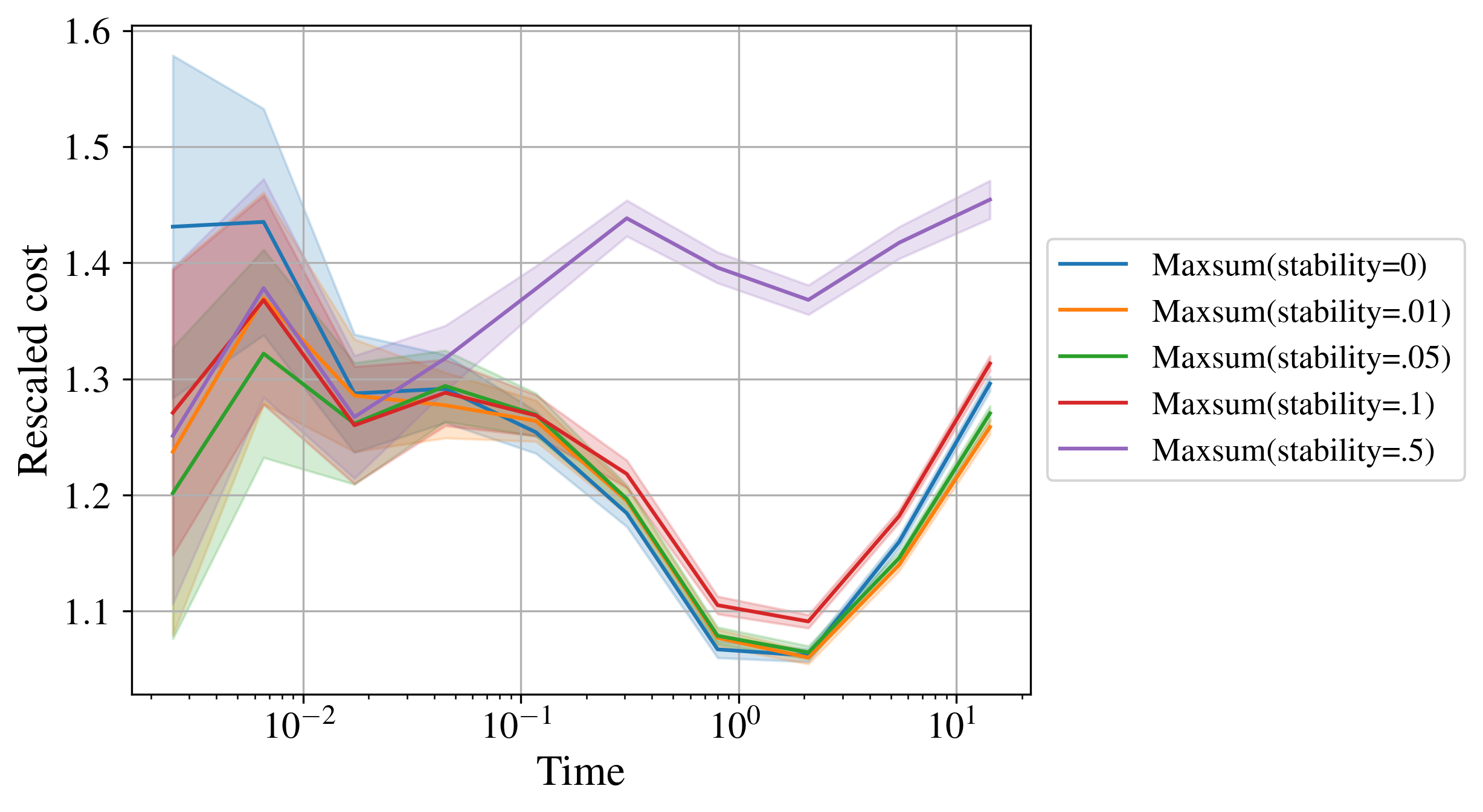}
    \caption{Cost of solution found over time for different stability values in Maxsum.}
    \label{fig:gc_hp_MS_stab}
\end{figure}

When comparing different levels of stability in the Maxsum algorithm, we find that using a stability parameter of $0.01$ performs the best (\Cref{fig:gc_hp_MS_stab}).
\subsubsection{MGM}
\begin{figure}[h!]
    \centering
    \includegraphics[width=\imgwidth]{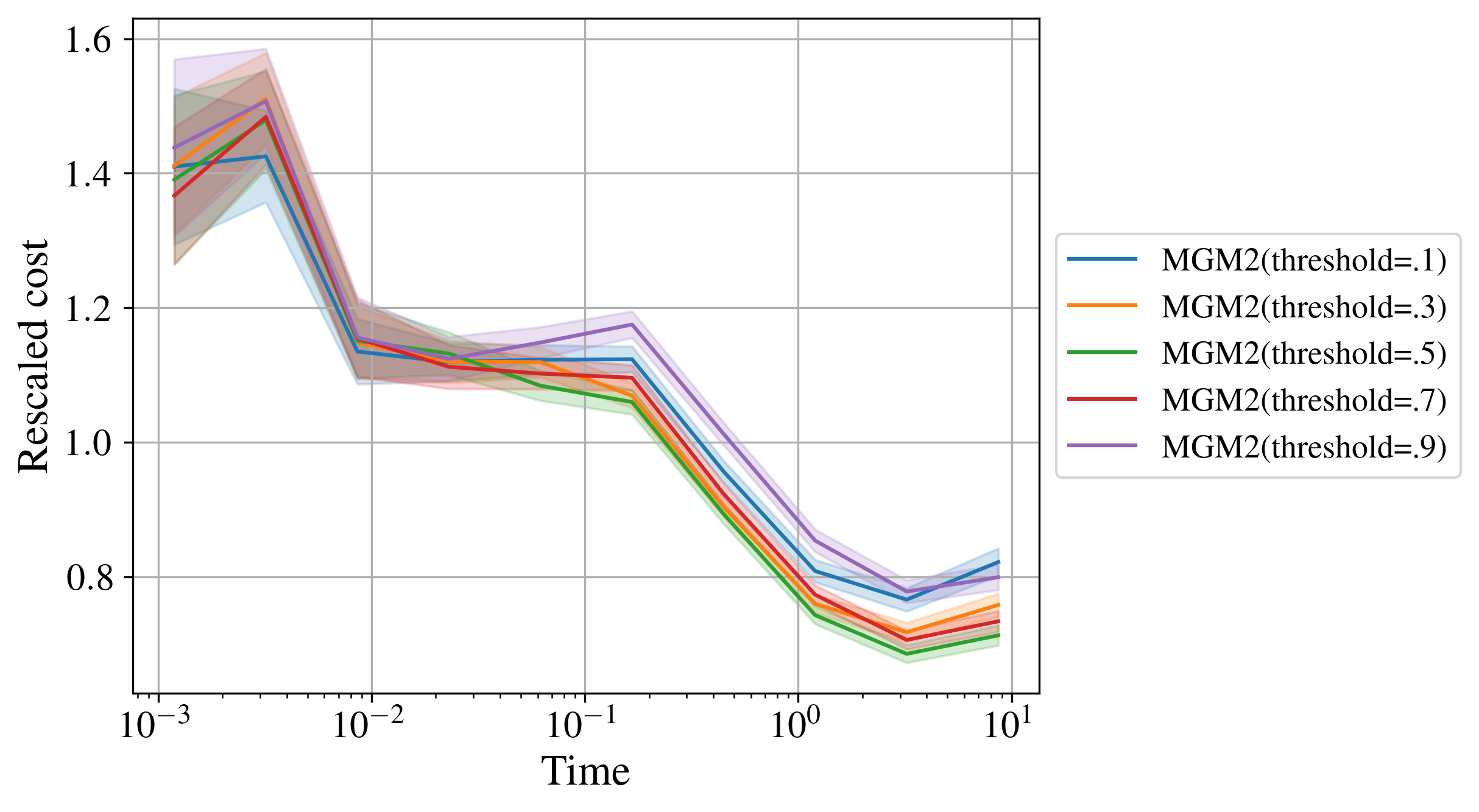}
    \caption{Cost of solution found over time for different threshold values in MGM2.}
    \label{fig:gc_hp_MGM}
\end{figure}

When comparing different values of threshold in the MGM2 algorithm, we find that using a stability parameter of $0.5$ performs the best (\Cref{fig:gc_hp_MGM}).



\subsection{Non context-based algorithm variants}
\label{sec:context-based-comparison}

\ifthenelse{\boolean{isSingleColumn}}{%
        \renewcommand{\imgwidth}{0.5\linewidth}
    }{%
        \renewcommand{\imgwidth}{0.7\linewidth}
    }%
\begin{figure}[htbp!]
    \centering
    \includegraphics[width=\imgwidth]{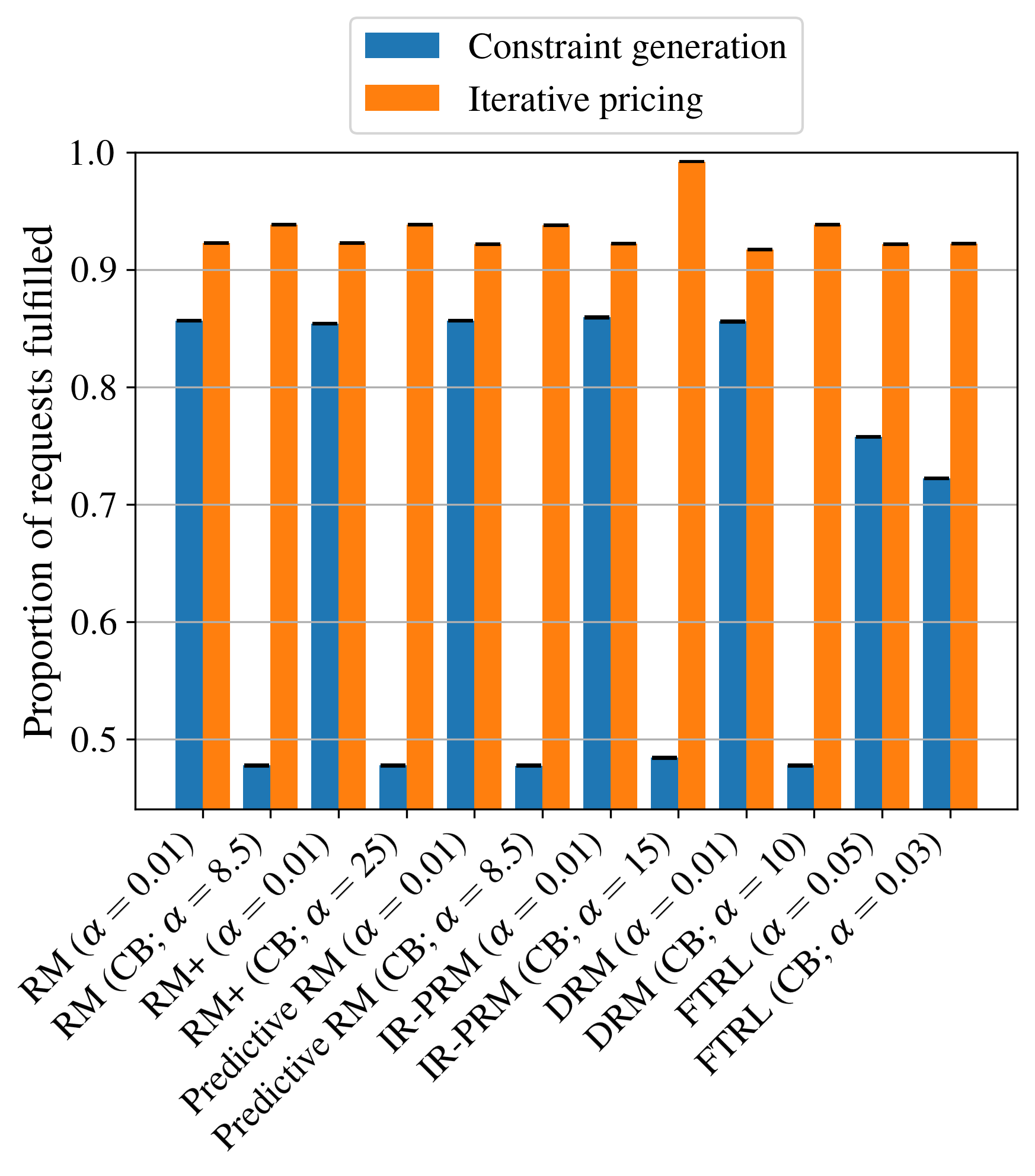}
    \caption{Utility achieved by standard and context-based (CB) variants of online learning algorithms. Standard error of the sample mean is displayed.}
    \label{fig:context-based-cmp}
\end{figure}
We compare the difference between the context-based and non context-based versions of each algorithm when used in large-scale COSPs.
For each algorithm variant, we set max iterations to 25 and tune the step size $\alpha$.
We take the average across 20 COSP problems, with 20 trials each.
We find that across all algorithms, using the context-based variant improves performance in the iterative pricing framework, and hurts performance in the constraint generation framework.
The best combination found is context-based IR-PRM used with iterative pricing, which fulfills 99.2\% of requests on average (Fig. \ref{fig:context-based-cmp}).
These results support our findings in the paper, where iterative pricing outperforms constraint generation when paired with any online learning algorithm.
To keep our plots clean, we display only the context-based variant of each algorithm in our main results.

\subsection{Step size for iterative pricing}
\label{sec:step-size-tuning}

\begin{figure*}[htbp!]
    \centering
    \subfigure[Default algorithm variants]{\includegraphics[width=.45\linewidth]{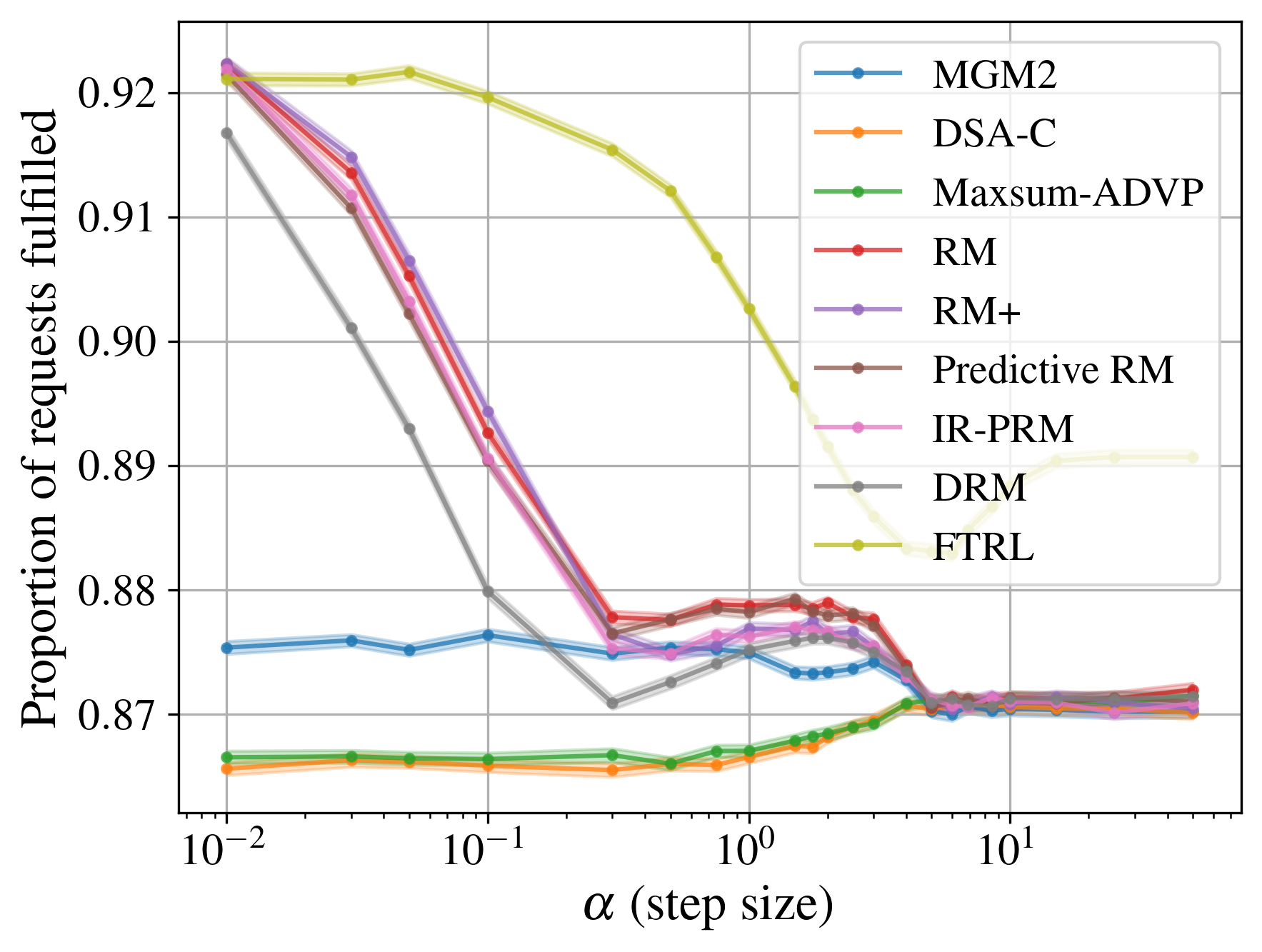}}
    \subfigure[Context-based variants]{\includegraphics[width=.45\linewidth]{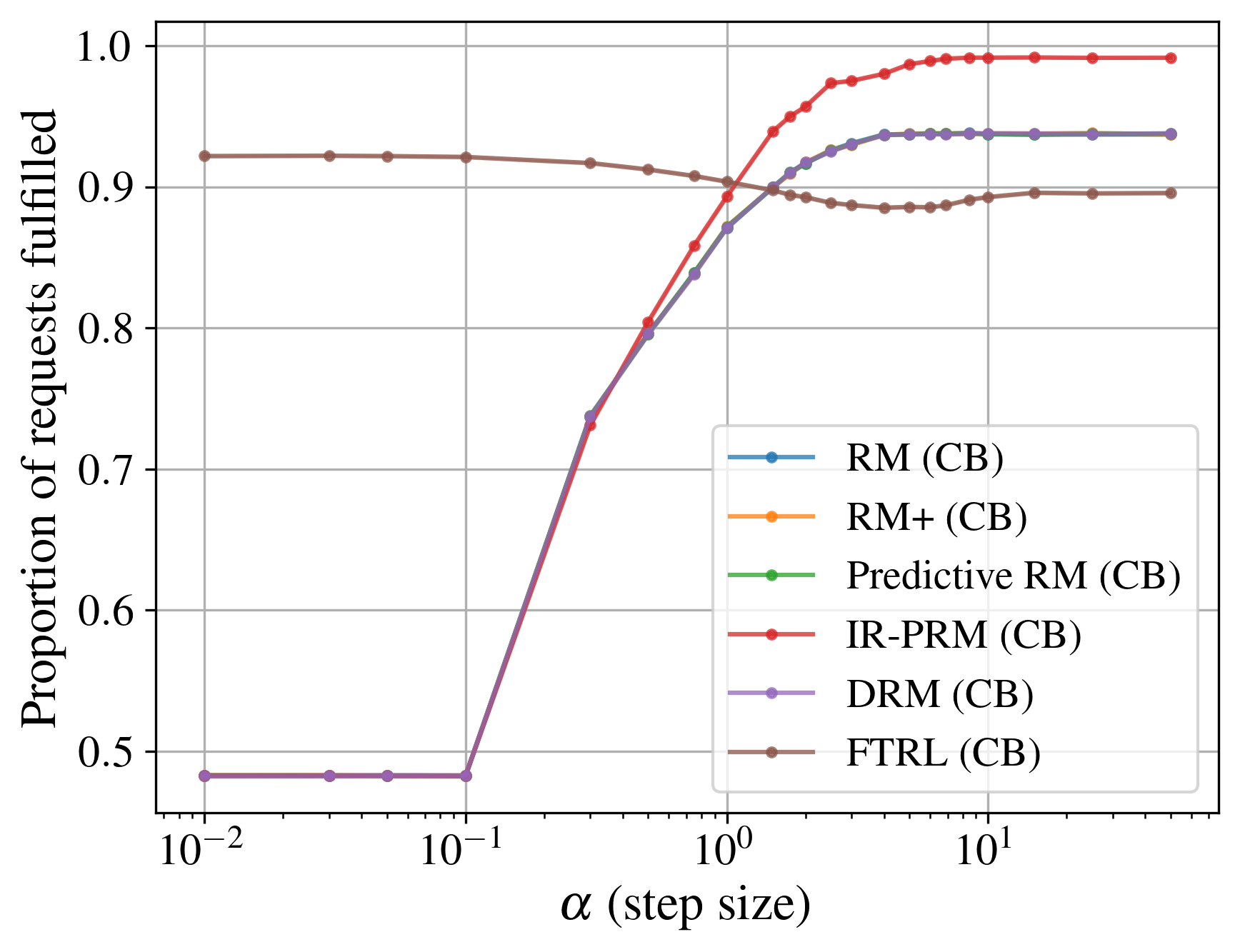}}
    \caption{Utility achieved by the iterative pricing framework using different values of $\alpha$. Standard error of the sample mean is displayed as the shaded region.}
    \label{fig:step-size-tuning}
\end{figure*}

While constraint generation is parameter free, iterative pricing requires us to set a step size (which we denote $\alpha$, and keep fixed throughout the algorithm).
Experimentally, we find that for iterative pricing in a large COSP, the optimal $\alpha$ depends on the algorithm used.
When inspecting the graph of performance for each $\alpha$ value (Fig. \ref{fig:step-size-tuning}), there appear to be four general shapes of curves.
\begin{itemize}
    \item All context-based online learning algorithms apart from FTRL improve performance as $\alpha$ gets large.
    
    \item Context-based FTRL and all non context-based variants of online learning algorithms appear to have a peak near $\alpha=0$, and a local optima at a larger value of $\alpha$ (around $\alpha=15$ for FTRL, and $\alpha=1.8$ for all others).
    
    \item Both DSA-C and Maxsum-ADBP have a slight boost in performance for $\alpha>3$ but are otherwise unaffected.

    \item MGM2 has a slight dip in performance for $\alpha>3$ and is otherwise unaffected.
    
\end{itemize}

In general, the $\alpha$ value chosen has drastic impact on the performances of most algorithms.
To account for this, we use the optimal alpha values (among the sweep that we performed) for each algorithm when doing our comparisons.

\subsection{Number of iterations}
\label{sec:max-iterations}

\begin{figure*}[htbp!]
    \centering
    \subfigure[Constraint generation]{\includegraphics[width=.45\linewidth]{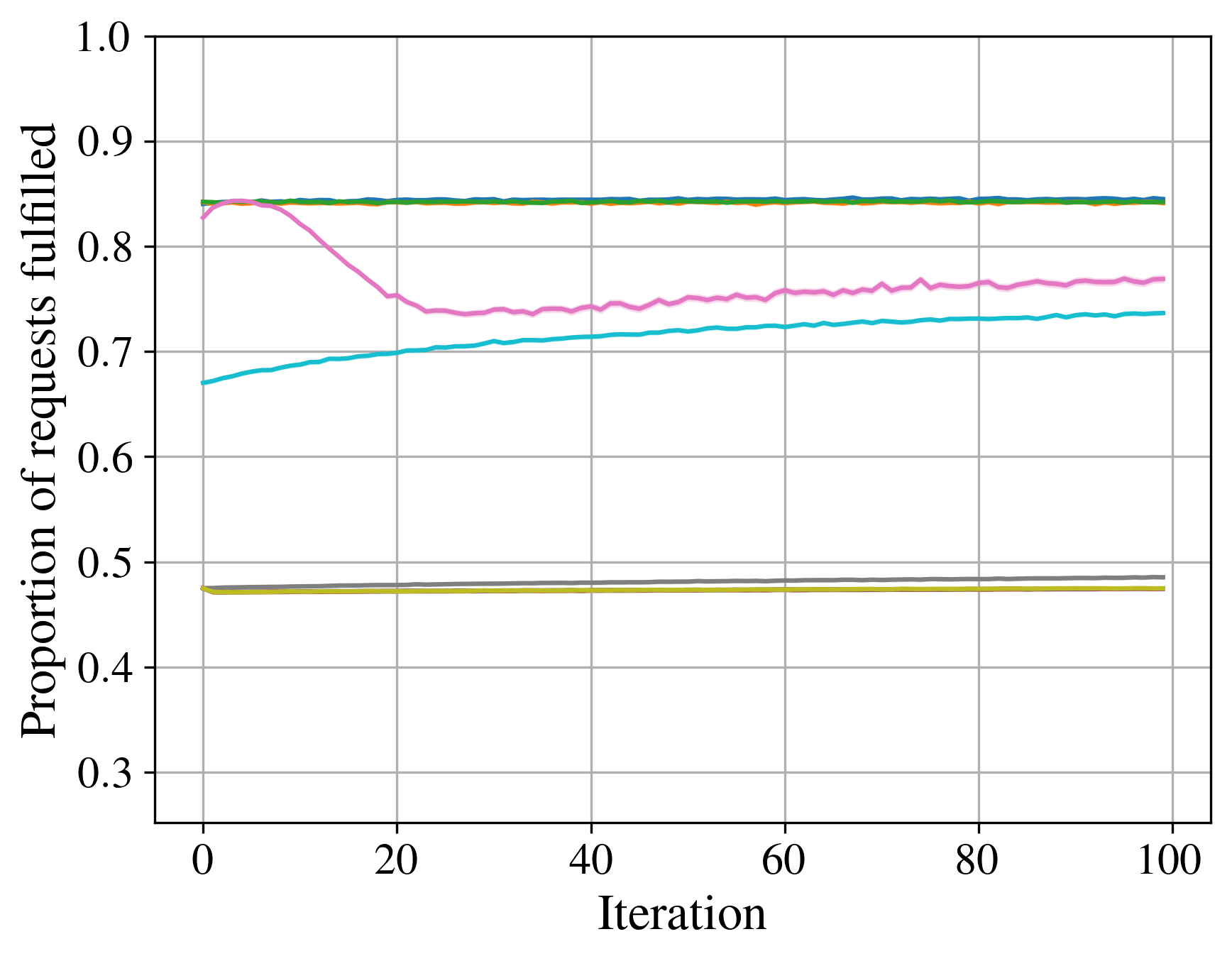}}
    \subfigure[Iterative Pricing]{\includegraphics[width=.45\linewidth]{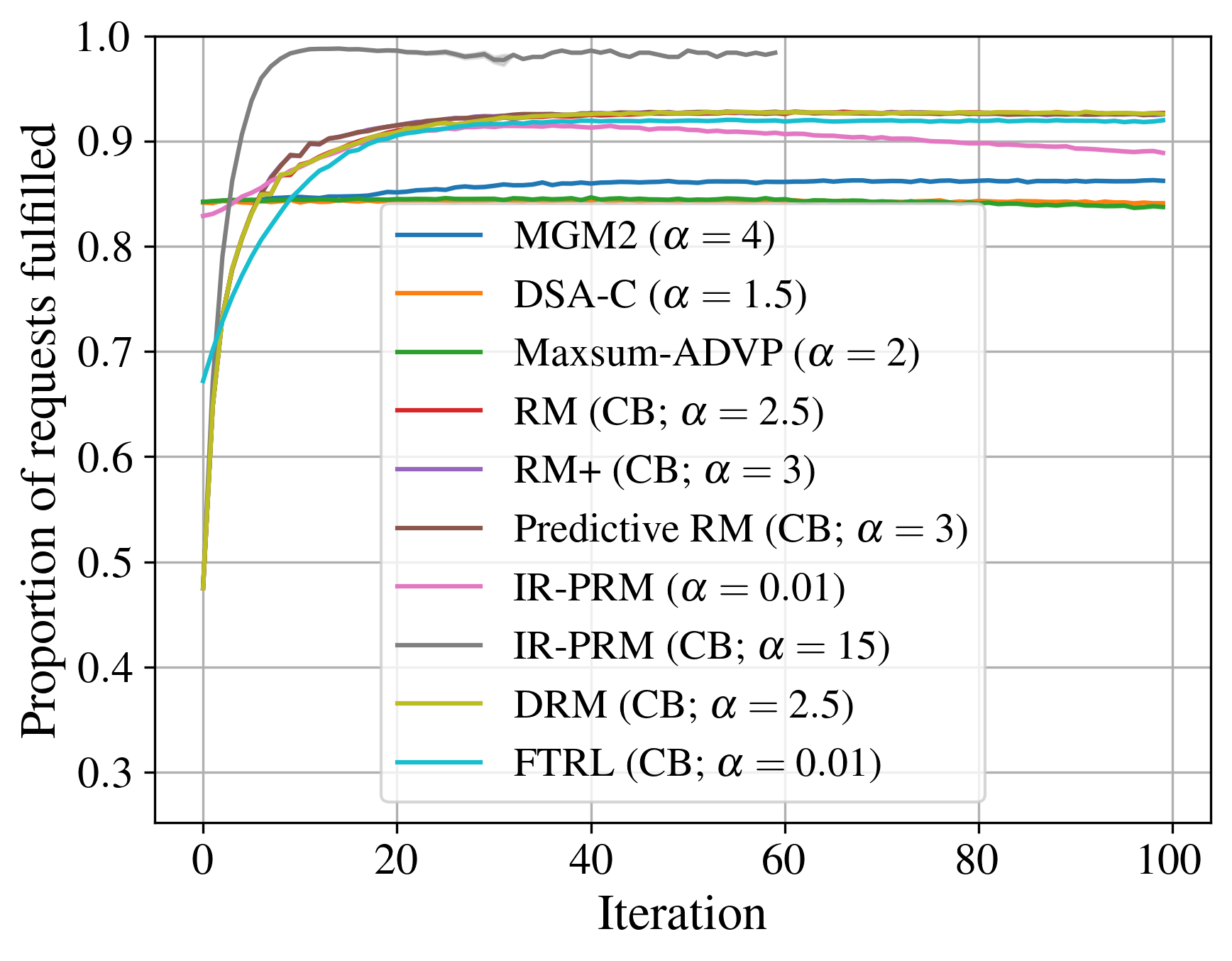}}
    \caption{Utility achieved per iteration. Standard error of the sample mean is displayed as the shaded region.}
    \label{fig:max-iterations}
\end{figure*}
To determine the setting of max iterations, we inspect the utility obtained per iteration.
When plotting up to iteration 100 with optimal $\alpha$ values\footnote{These $\alpha$ values are optimized for max iterations set to 100, so they are different from our main experiments.} (Fig. \ref{fig:max-iterations}), we find that most of our considered algorithms have achieved close to their optimal score by about iteration 25.
The only exception is context-based FTRL under constraint generation.
Since that particular counterexample fails to achieve high performance anyway, we use 25 as our maximum number of iterations in our experiments.
\section{Final hyperparameter settings}
In~\cref{tab:hyperparam-list-gc,tab:hyperparam-list-ss}, we list the final hyperparameters for each experiment in the paper.

\ifthenelse{\boolean{isSingleColumn}}{%
        \renewcommand{\tabwidth}{0.6\linewidth}
    }{%
        \renewcommand{\tabwidth}{\linewidth}
    }%
\begin{table}[h!]
    \centering
    \resizebox{\tabwidth}{!}{
    \begin{tabular}{lcc}
        \toprule
        \textbf{Algorithm} & \textbf{Hyperparameter} & \textbf{Value}\\
        \midrule
        \multirow{2}*{DRM} &  $\alpha$ (Positive discount) & 1.5 \\
         &  $\beta$ (Negative discount) & 0 \\\hline
        \multirow{1}*{DRM+} &  $\alpha$ (Positive discount) & 1.5 \\\hline
        \multirow{2}*{Damped DRM+} &  $\alpha$ (Positive discount) & 1.5 \\
         & Damping & 0.3 \\\hline
         \multirow{1}*{FTRL} &  $\eta$ (Regularization) & 1 \\\hline
         \multirow{2}*{Maxsum-ADVP} &  Damping & 0.8 \\
         &  Stability & 0.01 \\\hline
         \multirow{3}*{GDBA} &  Cost modification  & Additive \\
         & Increase mode & E \\
         & Constraint violation & Non-zero \\\hline
         \multirow{2}*{DSA-C} &  Variant & C \\
         & $p$ (Value change probability) & 0.5 \\\hline
         \multirow{2}*{MGM2} &  Threshold & 0.5 \\
         & Move favored & Unilateral \\
        \bottomrule
    \end{tabular}}
    \caption{Algorithm hyperparameters for graph coloring experiments.}
    \label{tab:hyperparam-list-gc}
\end{table}

\begin{table}[ht!]
    \centering
    \resizebox{\tabwidth}{!}{
    \begin{tabular}{lcc}
        \toprule
        \textbf{Algorithm} & \textbf{Hyperparameter} & \textbf{Value}\\
        \midrule
        \multirow{3}*{MGM2} &  Threshold & 0.5 \\
        & Move favored & Unilateral \\
        & $\alpha$ (Step size) & 0.1 \\\hline
        \multirow{3}*{DSA-C} &  Variant & C \\
        & $p$ (Value change probability) & 0.5 \\
        & $\alpha$ (Step size) & 8.5 \\\hline
        \multirow{3}*{Maxsum-ADVP} & Damping & 0.5 \\
        & Stability & 0.0001 \\
        & $\alpha$ (Step size) & 50 \\\hline
        \multirow{1}*{Context-based RM} & $\alpha$ (Step size) & 8.5 \\\hline
        \multirow{1}*{Context-based RM+} & $\alpha$ (Step size) & 25 \\\hline
        \multirow{1}*{Context-based Predictive RM} & $\alpha$ (Step size) & 8.5\\\hline
        \multirow{1}*{IR-PRM} & $\alpha$ (Step size) & 0.01 \\\hline
        \multirow{1}*{Context-based IR-PRM} & $\alpha$ (Step size) & 15 \\\hline
        \multirow{3}*{Context-based DRM} & $\alpha$ (Positive discount) & 1.5 \\
         & $\beta$ (Negative discount) & 0 \\
        & $\alpha$ (Step size) & 10 \\\hline
        \multirow{1}*{Context-based FTRL} & $\alpha$ (Step size) & 0.03 \\
        \bottomrule
    \end{tabular}}
    \caption{Algorithm hyperparameters for satellite scheduling experiments.}
    \label{tab:hyperparam-list-ss}
\end{table}

\section{Compute environment}
\label{sec:appendix-compute}

All experiments were conducted in Python 3.8.20 on a Linux cluster. Compute nodes were equipped with dual AMD EPYC 7252 CPUs (16 physical cores, 32 hardware threads) and 512 GB of RAM. The graph coloring experiments were implemented in PyDCOP, an open-source Python library for DCOPs. The satellite scheduling experiments used a custom Python evaluation framework. The local scheduling oracles leverage the OR-Tools library and typically executed in milliseconds. All experiments were conducted using unseeded random number generators.

\section{Omitted figures}
\label{sec:appendix-figures}

\ifthenelse{\boolean{isSingleColumn}}{%
    \renewcommand{\imgheight}{105 pt}
}{%
    \renewcommand{\imgheight}{110 pt}
}%

\begin{figure*}[htbp!]
  \centering
    \subfigure[Random networks]{\includegraphics[height=\imgheight]{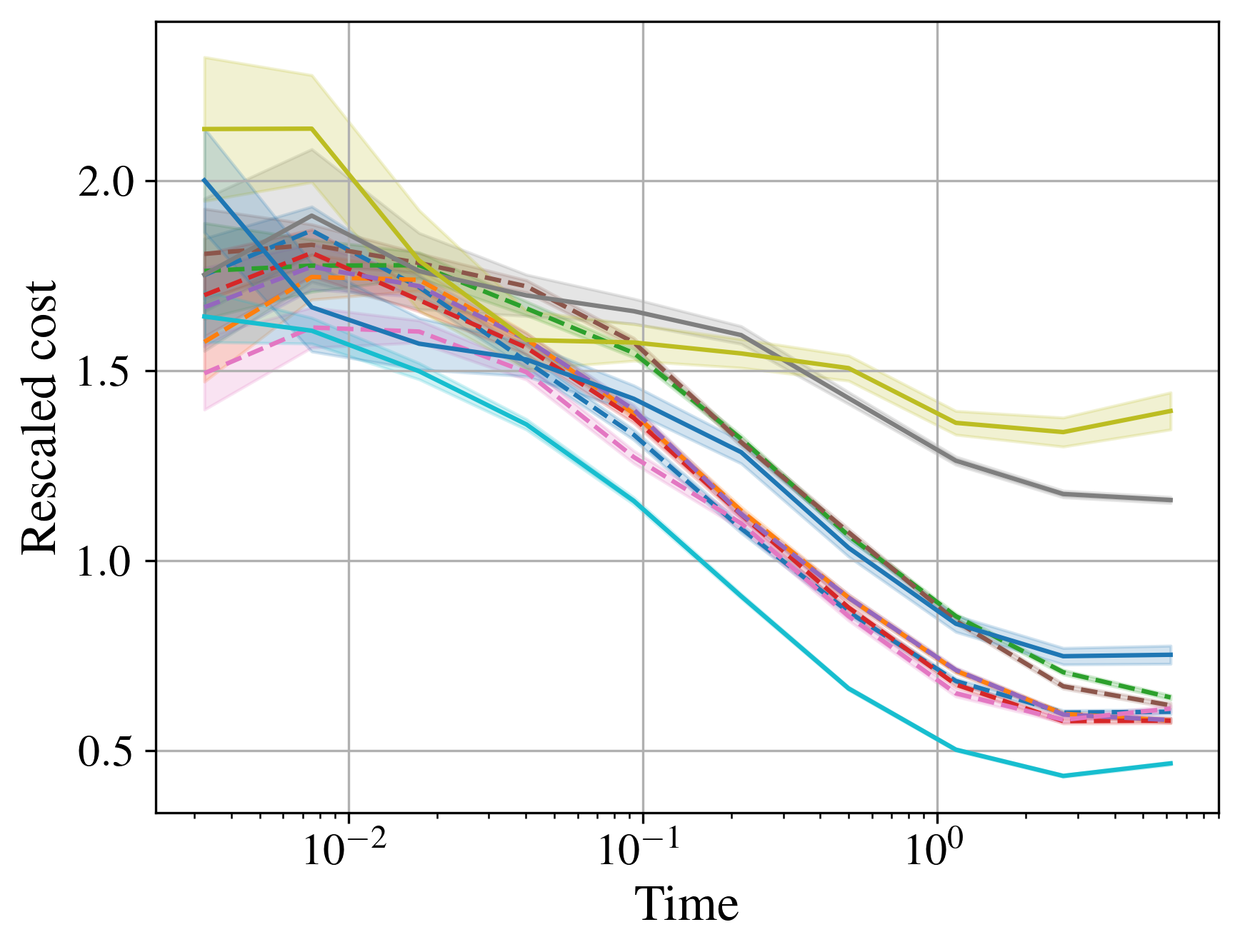}}
    \subfigure[Scalefree networks]{\includegraphics[height=\imgheight]{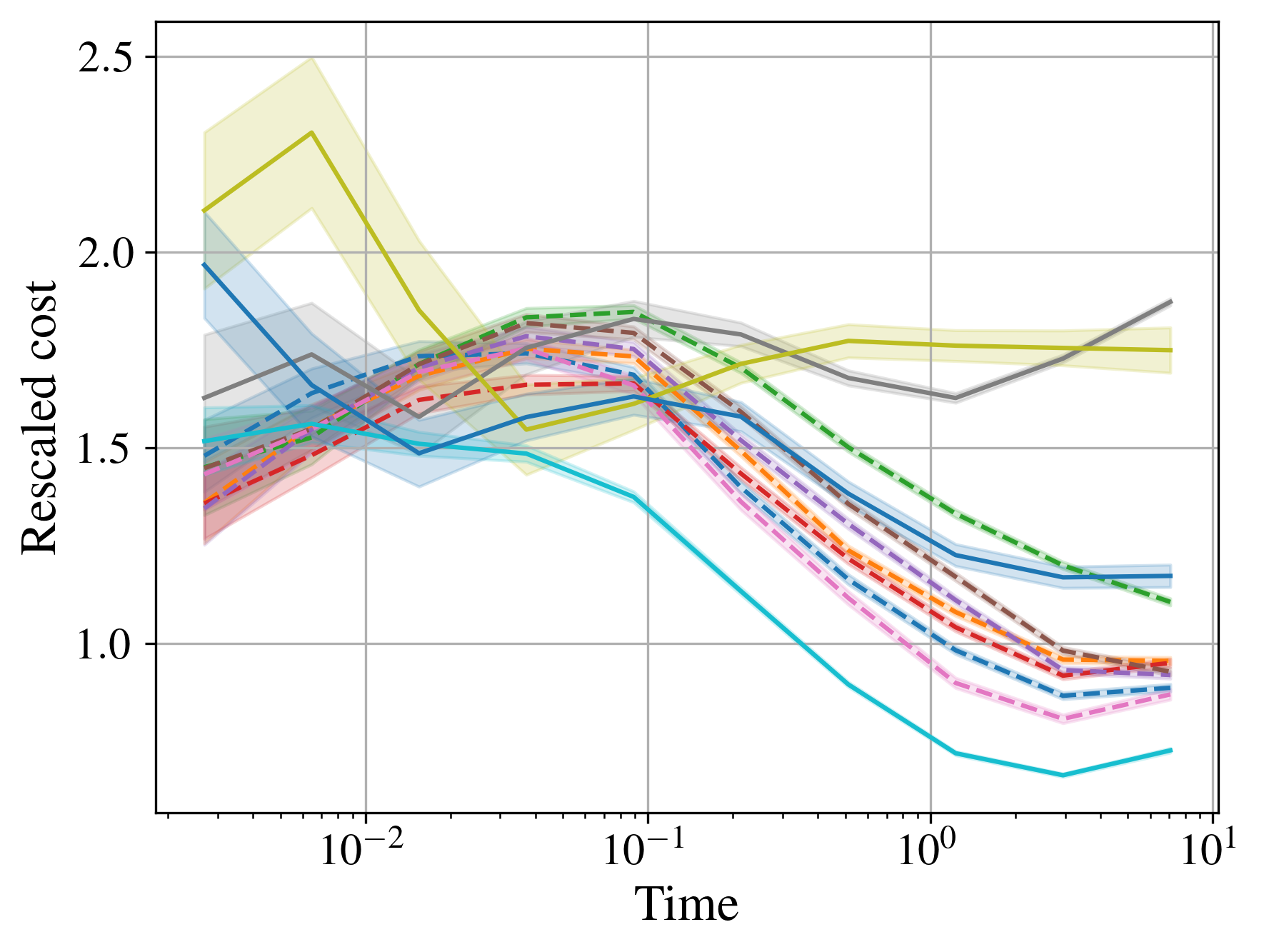}}
    \subfigure[All networks]{\includegraphics[height=\imgheight]{images/graph_coloring/cost_over_time/rescaled_combined_plot.png}}
    \caption{Cost of solution found over time for graph coloring problems by network type.}
    \label{fig:cost-over-time-appendix}
\end{figure*}

\ifthenelse{\boolean{isSingleColumn}}{%
    \renewcommand{\imgheight}{103 pt}
}{%
    \renewcommand{\imgheight}{110 pt}
}%
\begin{figure*}[htbp!]
    \centering
    \subfigure[$n=10$]{\includegraphics[height=\imgheight]{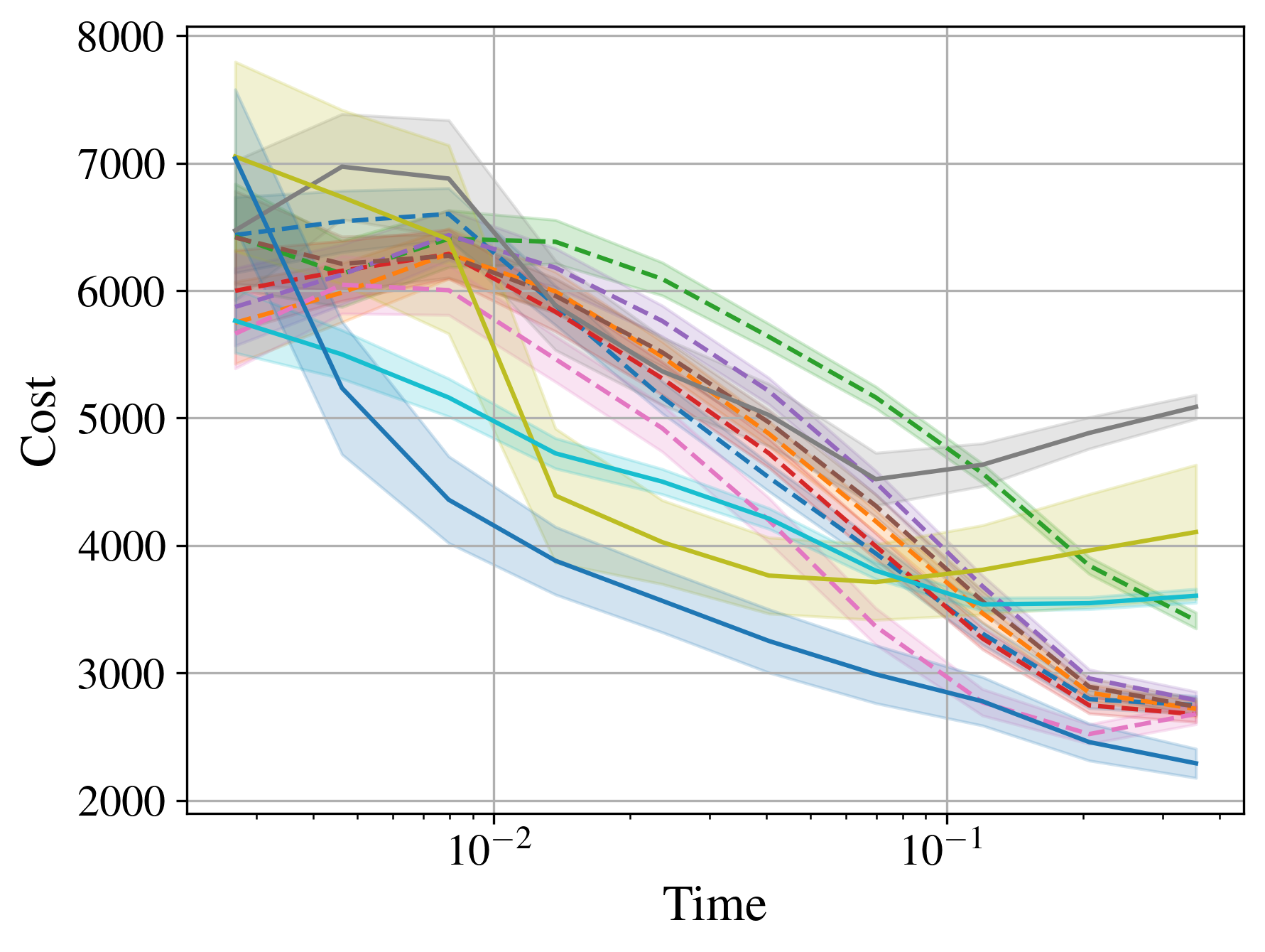}}
    \subfigure[$n=20$]{\includegraphics[height=\imgheight]{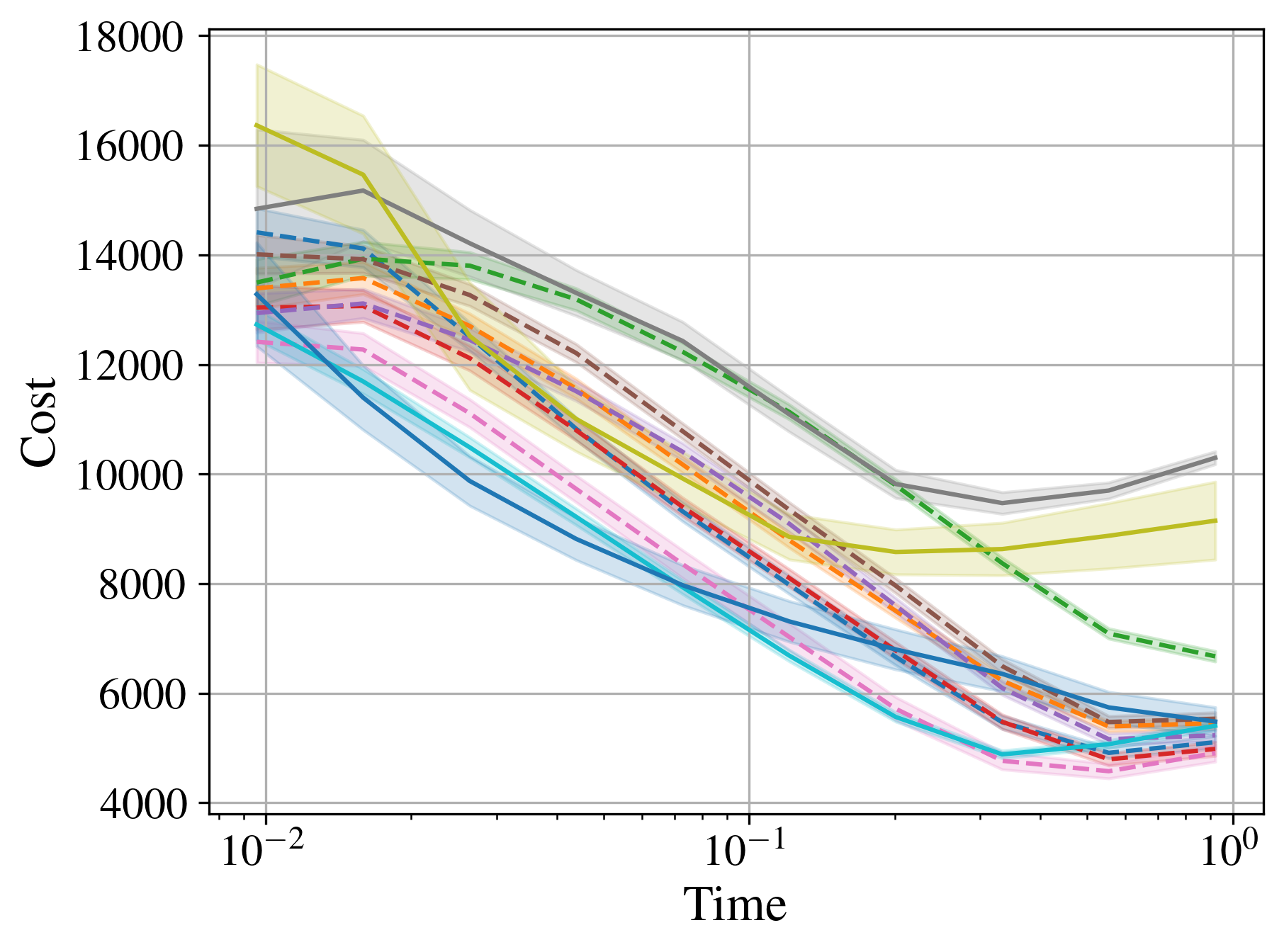}}
    \subfigure[$n=100$]{\includegraphics[height=\imgheight]{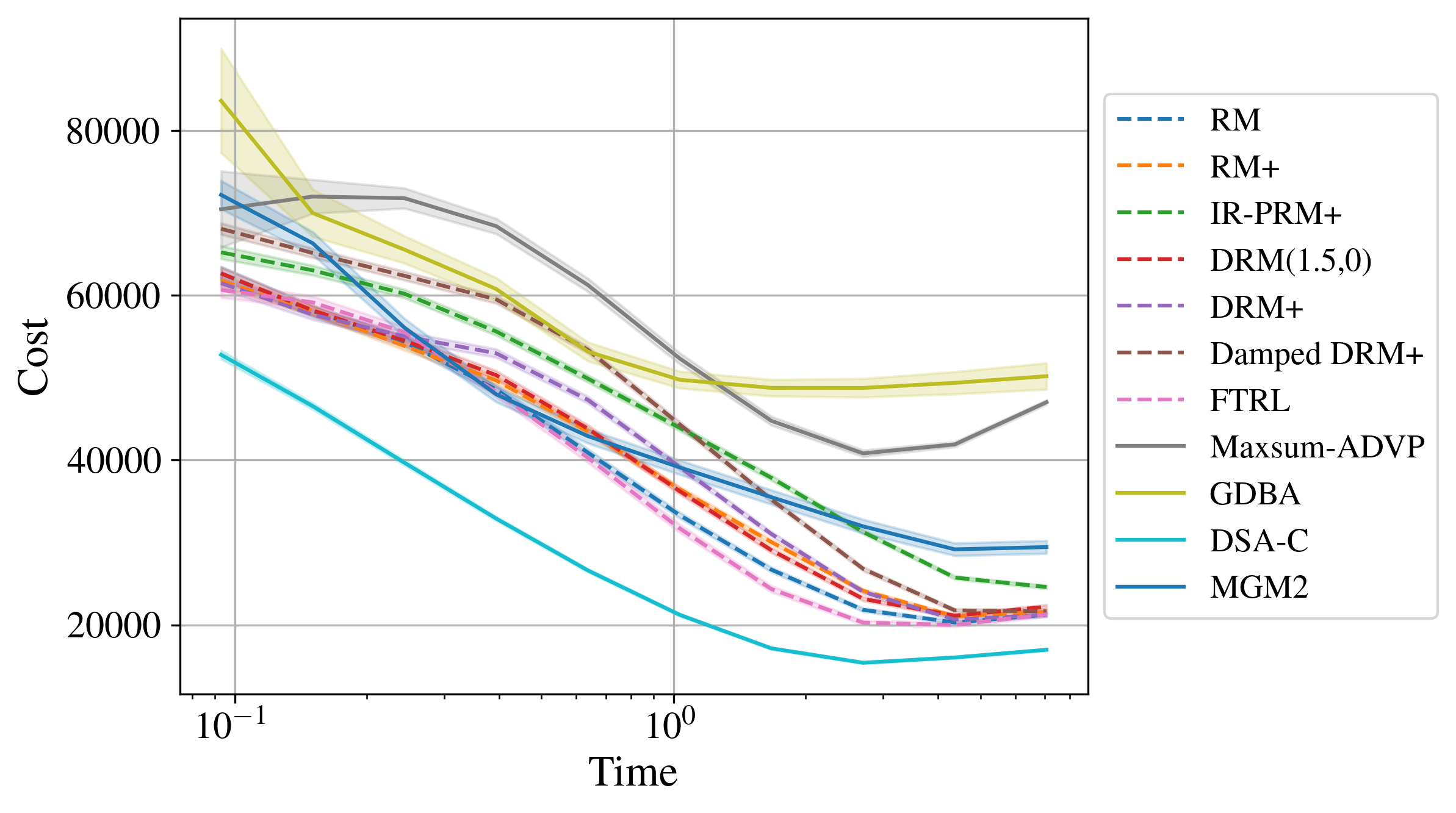}}
   \caption{Cost of solution found over time for graph coloring problems by network size.}
    \label{fig:cost-over-time-for-ns}
\end{figure*}

\end{document}